\documentclass{article}

\usepackage[preprint]{neurips_2026}
\usepackage[utf8]{inputenc}
\usepackage[T1]{fontenc}
\usepackage{hyperref}
\usepackage{url}
\usepackage{booktabs}
\usepackage{newtxtext}
\usepackage{amsfonts}
\usepackage{nicefrac}
\usepackage{microtype}
\usepackage{wrapfig}
\usepackage{xspace}
\usepackage{amsmath}
\usepackage{graphicx}
\usepackage{multirow}
\usepackage{threeparttable}
\usepackage[table]{xcolor}
\usepackage{subcaption}
\usepackage{makecell}
\usepackage[most]{tcolorbox}
\usepackage{amsthm}
\usepackage{soul}

\definecolor{rowgray}{HTML}{E8E8E8}
\definecolor{rowblue}{HTML}{EAF2F8}
\definecolor{rowpink}{HTML}{FCEFE3}
\definecolor{bestcolor}{HTML}{E66100}
\definecolor{secondcolor}{HTML}{2F7FBF}

\DeclareRobustCommand{\best}[1]{\textcolor{bestcolor}{\textbf{#1}}}
\DeclareRobustCommand{\second}[1]{\textcolor{secondcolor}{\underline{#1}}}

\newcommand{\icon}{\raisebox{-4pt}{\includegraphics[width=1.1em]{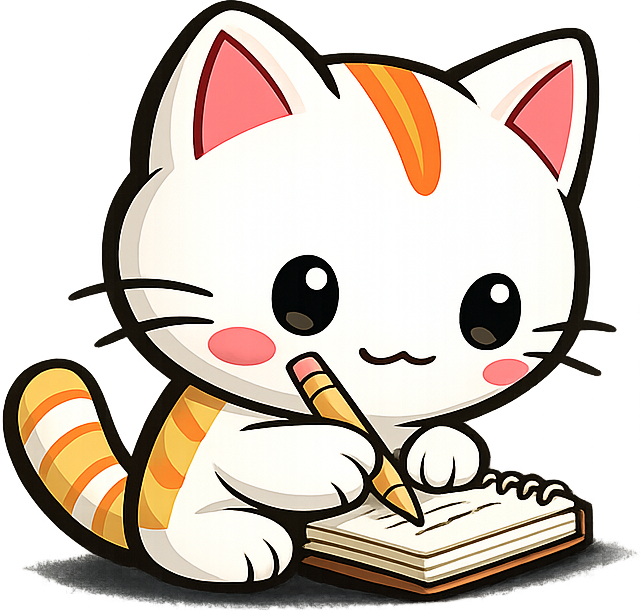}}\xspace}

\newcommand{\method}{\textsc{LeapTS}}

\title{\icon \method: Rethinking Time Series Forecasting as Adaptive Multi-Horizon Scheduling}

\author{
  Sheng Pan$^{1}$,\quad Ming Jin$^{1} \thanks{Corresponding authors.} $, \quad Bo Du$^{1}$, \quad Shirui Pan$^{1,*}$, \\
  $^1$Griffith University\quad\\
  \texttt{sheng.pan@griffith.uni.edu.au,}   \texttt{mingjinedu@gmail.com},  \\
  \texttt{bo.du@griffith.edu.au},
  \texttt{s.pan@griffith.edu.au}
}

\begin{document}

\maketitle

\begin{abstract}
Time series forecasting serves as an essential tool for many real-world applications, supporting tasks such as resource optimization and decision-making. Despite significant architectural advancements, most modern models still treat forecasting task as a fixed mapping from history to target horizons. This induces temporal decoupling across future time points and limits the model's ability to adapt to the evolving context as forecasting progresses. In this work, we present \method, a novel framework that reformulates time series forecasting as a \emph{dynamic scheduling process} over the prediction horizon. Specifically, \method\ organizes the forecasting process into multi-level decisions using: (1) the \emph{hierarchical controller} to dynamically select the optimal prediction scale and advancement length at each step, and (2) \emph{continuous-time state evolution} driven by neural controlled differential equations. Within this process, the \emph{controlled update mechanism} explicitly couples the irregular temporal dynamics with discrete scheduling feedback. Extensive evaluations on both real-world and synthetic datasets demonstrate that \method\ improves overall forecasting performance by at least 7.4\% while achieving a $2.6\times$ to $5.3\times$ inference speedup over representative Transformer-based models. Furthermore, by explicitly tracing the scheduling trajectories, we reveal how the model autonomously adapts its forecasting behavior to capture non-stationary dynamics. 
\end{abstract}

\section{Introduction}

Time series forecasting predicts future values from historical observations and plays an important role in many real-world systems, including energy planning \citep{hong2020energy}, demand forecasting \citep{carbonneau2008application}, health monitoring \citep{gul2009statistical}, resource scheduling \citep{hipel1994time}, and strategic decision-making \citep{sezer2020financial}. In recent years, many advanced deep models have achieved strong performance by learning rich historical representations \citep{liu2024itransformer,fei2025amplifier}. However, temporal patterns rarely evolve uniformly across the prediction horizon. Stable trends, abrupt fluctuations, and regime shifts can emerge at different future stages, making a fixed prediction process rigid and potentially brittle when confronting such dynamics.

\begin{figure*}[t]
    \centering
    \includegraphics[width=1\linewidth]{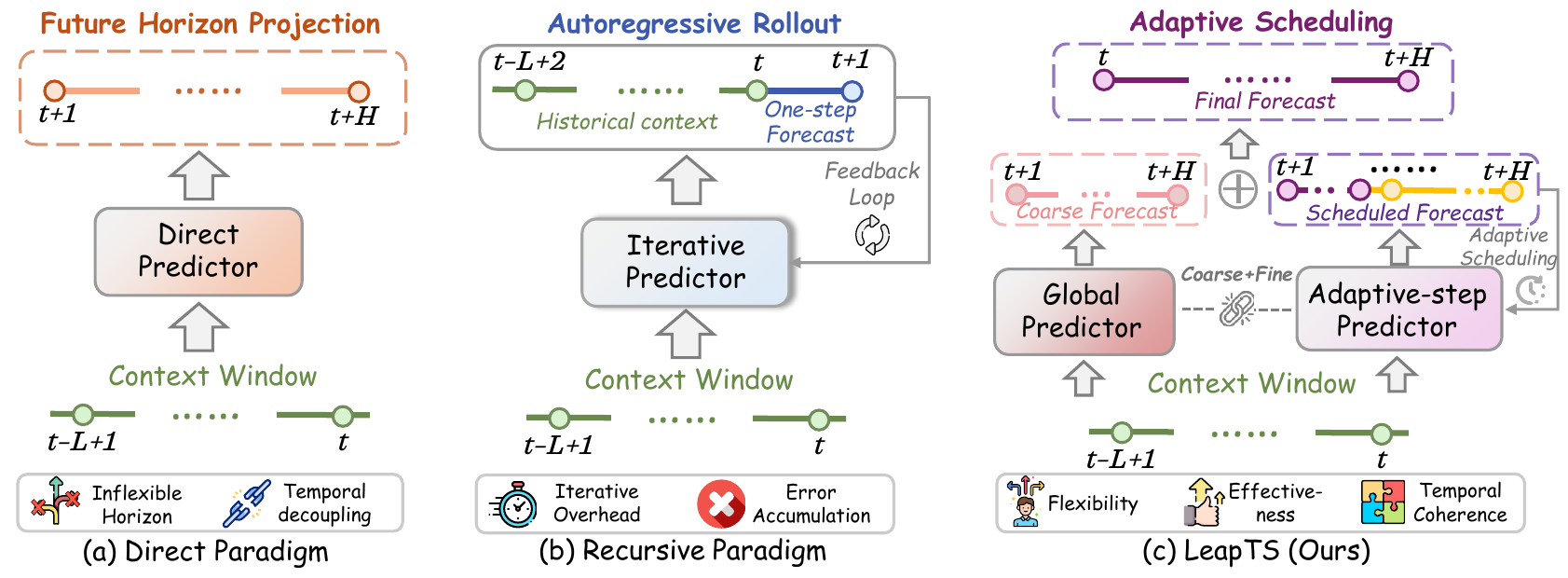}
    \caption{Comparison of multi-horizon forecasting paradigms. (a) The Direct paradigm predicts the full horizon in a single forward pass but is temporally decoupled. (b) The Recursive paradigm forecasts step by step but suffers from error accumulation. (c) Our paradigm formulates multi-horizon forecasting as a closed-loop adaptive scheduling process, enabling flexibility and temporal coherence.}
\label{fig:intro}
\vspace{-4mm}
\end{figure*}

Such issue largely stems from the execution paradigms used in existing multi-horizon forecasting methods. As illustrated in Fig.~\ref{fig:intro}(a-b), current methods mainly follow two paradigms. The \textit{recursive paradigm} generates future values step by step with a fixed stride, feeding previous predictions back into the model and thereby preserving temporal ordering \citep{beck2024xlstm}. However, it suffers from severe error accumulation and computational overhead over long horizons \citep{marcellino2006comparison}. As a result, recent deep architectures mostly adopt the \textit{direct paradigm}, which predicts the entire horizon in a single forward pass from historical representations \citep{taieb2012review}. While efficient, this design reduces forecasting to a fixed horizon-level mapping. This leads to three critical problems: \textbf{\textit{Overly complex encoders.}} Deep models increasingly rely on complex encoder designs to extract stronger historical representations, which may result in overfitting and reduce robustness \citep{wang2025accuracy,bergmeir2024fundamental}. \textbf{\textit{Lack of contextual adaptability.}} Although temporal patterns may vary substantially across future stages, models cannot update their internal states based on intermediate feedback. This makes them insensitive to dynamic transitions during the forecasting process \citep{kim2021reversible,liu2022non}. \textbf{\textit{Decoupled horizon dependencies.}} Future steps are naturally ordered, but the direct paradigm implicitly treats outputs as conditionally independent given the historical representation. This prevents models from capturing the temporal dependencies among them. Fundamentally, both paradigms suffer from structural rigidity: they enforce pre-defined execution paths that strictly decouple the forecasting process from the evolving temporal context. This raises a key question:

\begin{tcolorbox}[
  colback=gray!5!white,
  colframe=gray!60!black,
  width=0.95\columnwidth,
  boxrule=0.5pt,
  arc=1mm,
  auto outer arc,
  boxsep=0.7mm,
  top=1mm,
  bottom=1mm,
  center,
  before skip=10pt,
  after skip=10pt
]
\centering
\textcolor{black!85}{      
  \textit{\textbf{How to adaptively adjust prediction behavior based on varying contextual dynamics?}}
}
\end{tcolorbox}

Motivated by these insights, we propose \method, which reformulates forecasting as a \emph{coarse-to-fine dynamic scheduling process} (Fig.~\ref{fig:intro}(c)). Instead of using complex encoder-based architectures, \method\ couples a global coarse predictor with an adaptive scheduling branch, and organizes the prediction process into multi-level scheduling to dynamically select the optimal prediction scale and step length. Furthermore, to keep the controller state updated as scheduling progresses, \method\ incorporates two update mechanisms. The \textbf{intra-step update mechanism} compresses immediate forecasting results into feedback signals to guide the controller, while the neural controlled differential equation (NCDE) drives the \textbf{inter-step state evolution} to track continuous-time transitions between scheduling steps. With explicitly recorded scale choices, advancement lengths, and state updates, \method\ provides a traceable scheduling path for analyzing dominant signals and scheduling granularity across volatility regimes. Together, these designs enable \method\ to integrate adaptive step scheduling with feedback-driven state evolution within a unified framework. Our main contributions are summarized as follows:

\begin{itemize}
\item We rethink the paradigm in time series forecasting and reformulate it as a dynamic scheduling process. This perspective shifts the focus from increasingly complex representation learning designs to adaptive prediction strategies that evolve with contextual information.
\item We propose \method, a forecasting framework that dynamically adjusts prediction behavior according to evolving contextual information. Specifically, it incorporates a \textbf{H-Controller}, a hierarchical module for scale and step selection, together with a \textbf{controlled state update mechanism} to couple temporal dynamics, adaptive scheduling, and state evolution.
\item \method~provides valuable insights into prediction process through explicit scheduling mechanisms. We present visualization and analysis of the learned scheduling strategies, revealing how forecasting behaviors evolve during prediction and helping to alleviate the black-box nature of deep learning models.
\end{itemize}

\section{Related Work}

\subsection{Deep Forecasting Methods}
With the rapid evolution of deep learning, recent time series forecasting has been largely dominated by Transformer variants such as iTransformer \citep{liu2024itransformer}, PhaseFormer \citep{niu2025phaseformer}, and EMAformer \citep{zhang2026emaformer}. These models rely on self-attention mechanisms to learn representations from historical observations and capture long-range temporal dependencies. However, although self-attention is highly effective in NLP tasks, its permutation-invariant nature makes it less suitable for modeling temporal order in time series forecasting \citep{zeng2023transformers,li2023revisiting}. Beyond Transformers, recent studies improve representation learning through diverse designs, including multi-scale modeling \citep{wang2025timemixer++, hu2025adaptive}, informative patching strategies \citep{wu2026enhancing, liu2026rethinking}, and frequency-domain decomposition \citep{fei2025amplifier, yi2024filternet}. Despite achieving strong performance, these approaches remain fundamentally representation-centric. They mainly focus on improving representation learning from historical observations, with forecasting implemented by directly mapping encoded representations to future predictions.

\subsection{Multi-Horizon Forecasting Paradigms}
The implementation of multi-horizon forecasting has evolved across distinct model families. Among them, the \textit{recursive paradigm} is widely adopted by sequence modeling architectures. Classic autoregressive models like DeepAR \citep{salinas2020deepar}, as well as recent advanced state-space and recurrent networks such as Mamba \citep{gu2024mamba} and xLSTM \citep{beck2024xlstm}, fundamentally rely on this step-by-step generation design. However, this autoregressive formulation forces models to prioritize local transitions, often struggling to capture global structures over extended horizons \citep{zhou2021informer,bergsma2023sutranets}. In contrast, the \textit{direct paradigm} has become the standard for recent time series forecasting tasks. These methods employ fixed prediction heads to map encoded historical representations to all future horizons in a single forward pass \citep{wang2024deep,kim2025comprehensive}. While effectively bypassing error accumulation, this projection reduces forecasting to a static multi-output regression and fails to explicitly model transitions between intermediate future states \citep{wang2025fredf,wang2026time}. Recent combination paradigms attempt to integrate the advantages of both strategies \citep{green2025stratify}. Nevertheless, such pre-defined horizon partitioning imposes rigid constraints, which inherently restrict flexibility in adapting to diverse contextual dynamics. Different from all these paradigms, \method\ is the first to reformulate multi-horizon forecasting as an adaptive scheduling process. This enables a uniquely context-aware dynamic prediction strategy.

\section{Methodology}

\paragraph{Problem Definition} Given a multivariate time series $\mathbf{X}_{t-L+1:t}\in\mathbb{R}^{L\times N}$, \method\ aims to predict the future horizon $\mathbf{Y}_{t+1:t+P}\in\mathbb{R}^{P\times N}$, with $L$, $P$, and $N$ denoting the look-back length, prediction length, and number of variables, respectively. Formally, the forecasting model learns a mapping $\hat{\mathbf{Y}}_{t+1:t+P}=\mathcal{F}_{\Theta}\!\left(\mathbf{X}_{t-L+1:t}\right)$, where $\Theta$ denotes all learnable parameters of the model. Unless otherwise specified, we omit the batch index for clarity. 

\subsection{Overview of \method}

Fig.~\ref{fig:framework} illustrates the comprehensive pipeline of \method, which primarily comprises two interconnected components. Specifically, Fig.~\ref{fig:framework}(a) depicts the \emph{overall forecasting framework} and shows the end-to-end flow from historical observations to the final predictions. Fig.~\ref{fig:framework}(b) provides a detailed view of the \textsc{H-Controller}, which serves as the core scheduling unit. Overall, \method\ achieves strong performance while maintaining efficiency and scheduling traceability.

\noindent\textbf{Overall Framework.} \method\ first encodes the historical sequence $\mathbf{X}_{t-L+1:t}$ into initial latent states $\mathbf{Z}_0=\Psi_{\mathrm{enc}}\!\left(\mathbf{X}_{t-L+1:t}\right)$ via an MLP encoder $\Psi_{\mathrm{enc}}$. A coarse forecast is directly derived from $\mathbf{Z}_0$ through a linear projection $\hat{\mathbf{Y}}^{\mathrm{coarse}}=\mathcal{H}_{\mathrm{coarse}}(\mathbf{Z}_0)$. The core forecasting process is then performed by the scheduling branch. At each scheduling step $k$, the hierarchical controller maintains the multivariate controller state $\mathbf{H}_k=[\mathbf{h}_{k,1},\dots,\mathbf{h}_{k,N}]^\top\in\mathbb{R}^{N\times d_h}$ and computes variable-wise high-level scale distributions $\boldsymbol{\Pi}_k$ and low-level advancement lengths $\boldsymbol{\ell}_k$ (Section~\ref{HC}):

\begin{figure*}[t]
    \centering
    \includegraphics[width=1\linewidth]{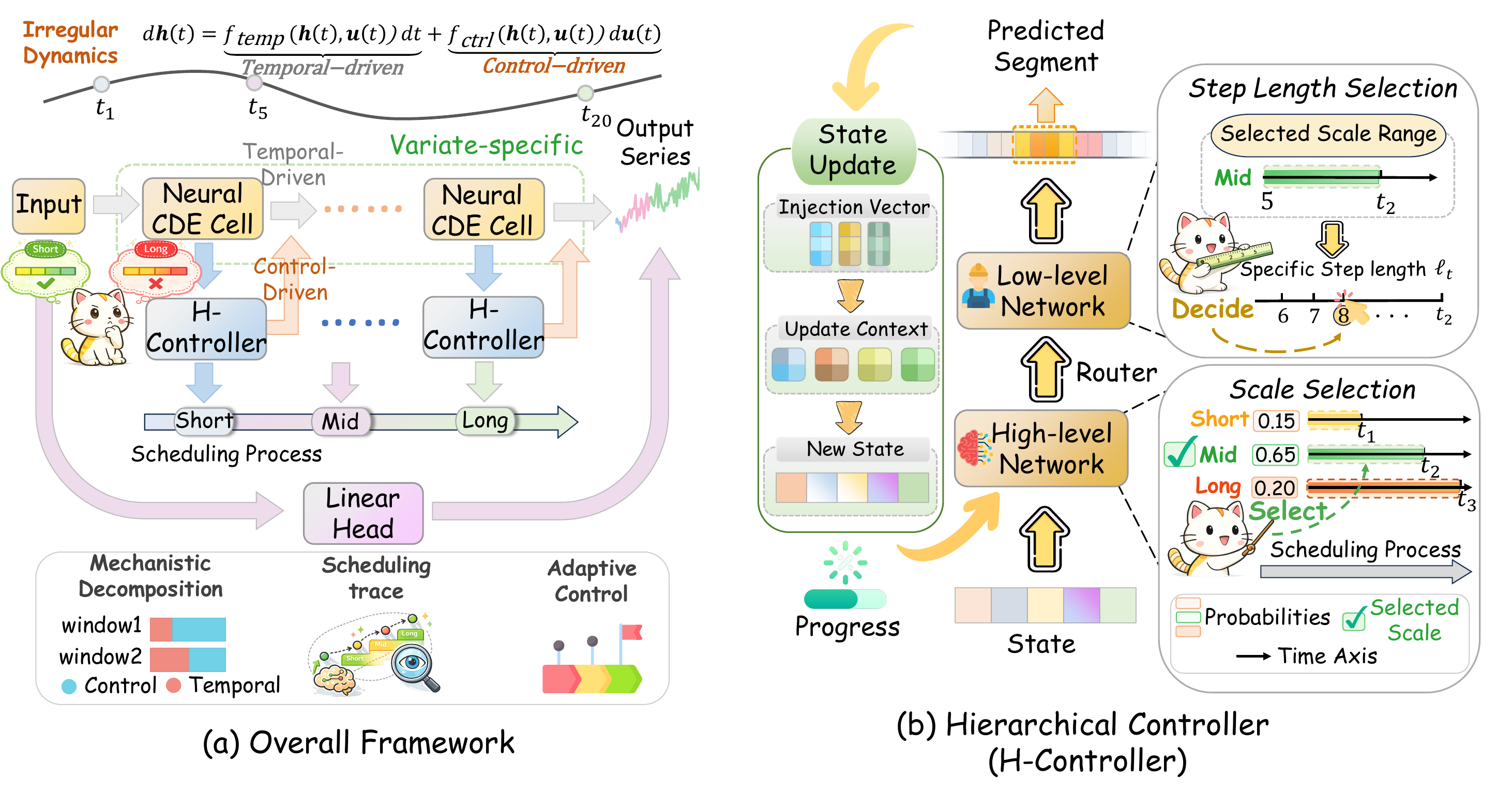}
    \caption{Framework of \method. (a) \method~reformulates multi-horizon forecasting as an adaptive scheduling problem. The scheduling branch progressively generates fine-grained forecasts and latent states evolve through variate-specific controlled dynamics. (b) The \textsc{H-Controller} performs scale selection then advancement-length selection. The generated segment is written into the target horizon, and the controller state updates based on current scheduling progress for subsequent steps.}
    \label{fig:framework}
    \vspace{-8mm}
\end{figure*}

\begin{equation}
\boldsymbol{\Pi}_k=\operatorname{HighLevel}(\mathbf{H}_k),
\quad
\boldsymbol{\ell}_k=\operatorname{LowLevel}(\mathbf{H}_k,\boldsymbol{\Pi}_k).
\end{equation}

Guided by these decisions, category-specific heads output the current forecast segment and progressively write it into the target horizon. Subsequently, the controller state is updated (Section~\ref{sue}) driven by the control signal $\mathbf{U}_k$, the control increment $\Delta\mathbf{U}_k$, and the temporal increment $\Delta\boldsymbol{\tau}_k$:
\begin{equation}
\mathbf{H}_{k+1}=\Phi_{\mathrm{evo}}(\mathbf{H}_k,\mathbf{U}_k,\Delta\mathbf{U}_k,\Delta\boldsymbol{\tau}_k).
\end{equation}
Finally, to integrate the coarse forecasts with fine-grained scheduling refinements, the two forecasts are fused by a learned gate with $\alpha\in(0,1)$: $\hat{\mathbf{Y}}=\hat{\mathbf{Y}}^{\mathrm{coarse}}+\alpha\hat{\mathbf{Y}}^{\mathrm{sched}}$. For notation simplicity, the remaining sections describe the univariate process. In the multivariate setting, the same formulation is applied to each variate.

\subsection{Hierarchical Controller} \label{HC}

The \textsc{H-Controller} serves as the core module in the scheduling branch. Instead of generating the entire forecasting horizon at once, it decomposes the iterative forecasting process into a sequence of hierarchical scheduling steps. At each scheduling step $k$, the controller first determines \emph{which scale to use} and then predicts \emph{how far the current step should advance}.

\noindent\textbf{High-Level Scheduler.}
Let $\mathbf{h}_k\in\mathbb{R}^{d_h}$ denote the univariate controller state at step $k$. The high-level scheduler produces a scale-selection distribution over three predefined scheduling categories, corresponding to short-, mid-, and long-range choices:
\begin{equation}
\boldsymbol{\pi}_k
=
\operatorname{Gumbel\text{-}Softmax}\!\left(\mathbf{W}_{\mathrm{cat}}\mathbf{h}_k\right)
\in \mathbb{R}^3,
\label{eq:schedule_category}
\end{equation}
where $\mathbf{W}_{\mathrm{cat}}$ is the category projection matrix. To enable end-to-end differentiability: the forward pass uses a hard one-hot selection to execute the discrete routing, while the backward pass propagates gradients through the continuous soft distribution (see Appendix~\ref{appendx:differentiability}). 

\noindent\textbf{Low-Level Scheduler.}  \label{step_section}
After the high-level scheduler determines the current scale, the low-level scheduler predicts the corresponding advancement length within the valid range of the selected scale. The detailed length ranges for different scales are provided in Appendix~\ref{appendx:length_ranges}. Specifically, for the selected category $c$, the advancement length is predicted as
\begin{equation}
\ell_{k,c}
=
L_c^{\min}
+
\bigl(L_c^{\max}-L_c^{\min}\bigr)\cdot
\sigma\!\left(f^{\mathrm{len}}_c(\mathbf{h}_k)\right),
\qquad c\in\{\mathrm{short},\mathrm{mid},\mathrm{long}\},
\label{eq:schedule_length_candidates}
\end{equation}

where $L_c^{\min}$ and $L_c^{\max}$ denote the minimum and maximum valid scheduling lengths for category $c$, and $f^{\mathrm{len}}_c(\cdot)$ denotes the category-specific length head, and $\sigma(\cdot)$ is the sigmoid activation function. Since the current scheduling category is one-hot in the forward pass, the executed advancement length is determined by the selected category:
\begin{equation}
\ell_k=\ell_{k,c^\ast},
\qquad
c^\ast=\arg\max_{c}\,\pi_{k,c}.
\label{eq:scheduled_length}
\end{equation}

\subsection{State Update and Evolution} \label{sue}

Once the decisions are made, the scheduling branch updates the forecasting state at two levels. The first is an intra-step summary-vector update within the \textsc{H-Controller}, which compresses the immediate scheduling outcomes. The second is the inter-step state evolution between consecutive scheduling steps, which propagates the scheduling state according to the current forecasting progress.

\noindent\textbf{Intra-Step Summary Update} Based on the selected scheduling category $c^\ast$ and the corresponding advancement length $\ell_k$, the current forecast segment is generated by the category-specific head:
\begin{equation}
\mathbf{s}_k=f^{\mathrm{seg}}_{c^\ast}(\mathbf{h}_k)\in\mathbb{R}^{P}.
\label{eq:selected_segment}
\end{equation}

Let $q_k$ denote the current cursor position on the prediction horizon, which updates via $q_{k+1} = q_k + \ell_k$. To write the selected segment according to the advancement length $\ell_k$ while maintaining differentiability, we introduce a continuous soft mask (see Appendix~\ref{appendx:differentiability}). Specifically, for each future time step $\tau \in \{1,\dots,P\}$, the mask value $m_k(\tau)$, the segment $\bar{\mathbf{s}}_k$, and the accumulated forecast $\hat{\mathbf{y}}^{\mathrm{sched}}_{k+1}$ are formulated as:
\begin{equation}
m_k(\tau)
=
\mathbb{I}[\tau\ge q_k]\cdot
\sigma\!\left(
\frac{\ell_k-(\tau-q_k)-0.5}{\gamma}
\right),
\label{eq:soft_mask}
\end{equation}
\begin{equation}
\bar{\mathbf{s}}_k = \mathbf{s}_k\odot \mathbf{m}_k,
\label{eq:effective_segment}
\end{equation}\vspace{-2mm}
\begin{equation}
\hat{\mathbf{y}}^{\mathrm{sched}}_{k+1}
=
\hat{\mathbf{y}}^{\mathrm{sched}}_{k}
+
\bar{\mathbf{s}}_k,
\qquad
\hat{\mathbf{y}}^{\mathrm{sched}}_{1}=\mathbf{0},
\label{eq:iterative_update}
\end{equation}

where $\mathbf{m}_k=[m_k(1),\dots,m_k(P)]^\top$, $\bar{\mathbf{s}}_k$ denotes the scheduled segment, and $\gamma > 0$ is a temperature controlling the mask sharpness. This formulation relaxes non-differentiable hard truncation into a smooth gate: the indicator $\mathbb{I}[\tau\ge q_k]$ prevents backward overwriting, while the $-0.5$ offset precisely centers the decision boundary between discrete steps to prevent information leakage. The segment $\bar{\mathbf{s}}_k$ is then compressed into a feedback signal:
\begin{equation}
\mathbf{c}_k=\tanh\!\left(\mathbf{W}_{\mathrm{sum}}\bar{\mathbf{s}}_k\right),
\label{eq:summary_signal}
\end{equation}
where $\mathbf{W}_{\mathrm{sum}}$ is the summary projection matrix. The resulting summary vector compresses the current outcome into a compact representation and is used by the \textsc{H-Controller} at the next step.

\noindent\textbf{Inter-Step State Evolution.} 
Between steps, the controller state evolves irregularly with scheduling progress. To support this, the current scheduling context is first encoded into a control signal:
\begin{equation}
\mathbf{u}_k
=
\tanh\!\left(
\mathbf{W}_{u}
\left[
\rho_k
\;\Vert\;
\bar{\ell}_{k-1}
\;\Vert\;
\boldsymbol{\pi}_{k-1}
\;\Vert\;
\mathbf{c}_{k-1}
\right]
\right),
\label{eq:controller_input}
\end{equation}
where $\rho_k$ denotes the remaining-horizon ratio, $\bar{\ell}_{k-1}$ denotes the normalized advancement length from the previous step, $\Vert$ denotes vector concatenation, and $\mathbf{W}_{u}$ is the control projection matrix. Based on this signal, the control and temporal increments are defined as:
\begin{equation}
\Delta\mathbf{u}_k=\mathbf{u}_k-\mathbf{u}_{k-1},
\label{eq:control_increment}
\end{equation}
\begin{equation}
\Delta\tau_k=\operatorname{clip}(\bar{\ell}_{k-1},\tau_{\min},\tau_{\max}),
\label{eq:temporal_increment}
\end{equation}
where $\Delta\mathbf{u}_k$ captures the variation of the control signal and $\Delta\tau_k$ denotes the temporal increment induced by the current scheduling progress. To exploit similar temporal patterns, the variates are grouped into clusters $g \in \{1,\dots,G\}$ via $\mathcal{L}_2$ distance minimization, so that variates within the same cluster share the corresponding evolution dynamics. The controller state is then evolved by the cluster-specific NCDE transition:
\begin{equation}
d\mathbf{h}(t)
=
\mathbf{F}_{g}\!\left(\mathbf{h}(t),\mathbf{u}(t)\right)\,d\mathbf{u}(t)
+
\mathbf{G}_{g}\!\left(\mathbf{h}(t),\mathbf{u}(t)\right)\,dt,
\label{eq:ncde_continuous}
\end{equation}

where $\mathbf{F}_{g}(\cdot)$ and $\mathbf{G}_{g}(\cdot)$ denote the cluster-specific vector fields. During scheduling, the state evolution in Eq.~\eqref{eq:ncde_continuous} is directly driven by control ($\Delta\mathbf{u}_k$) and temporal ($\Delta\tau_k$) increments. This explicitly couples the irregular scheduling steps with the hidden state update, while preserving parameter sharing within each variable cluster. Through repeated intra-step summary updates and inter-step evolution, the iterative branch progressively refines the forecasting state.

\section{Experiments}

We conduct extensive experiments to evaluate \method. Section~\ref{sec:real_forecasting} reports results on real-world datasets. Section~\ref{sec:sim_forecasting} evaluates the model on complex synthetic scenarios. Section~\ref{sec:explainability} presents analyses to reveal internal behaviors. Section~\ref{sec:ad_ana} provides efficiency and ablation studies. \textbf{In summary, \method\ achieves strong forecasting performance while providing valuable insights into its scheduling dynamics.} Detailed experimental configurations and implementations are in Appendix~\ref{appendx:imp_detail}.

\subsection{Performance on Real-World Datasets}
\label{sec:real_forecasting}

\subsubsection{Multivariate Forecasting}
\label{subsec:real_multi}

\noindent\textbf{Setups.} We evaluate our model on 13 widely-used real-world datasets across diverse domains: \textbf{Energy} (ETTh1/2, ETTm1/2, Electricity), \textbf{Weather}, \textbf{Traffic} (PEMS04/07), \textbf{Finance} (Exchange, SP500), \textbf{Healthcare} (ILI, COVID-19), and \textbf{Web} (Wiki). The input length for all models is fixed at 96 (36 for the ILI dataset), and we evaluate performance across multiple horizons $P \in \{18, 24, 36, 48, 60\}$. Mean Squared Error (MSE) and Mean Absolute Error (MAE) are used to assess forecasting accuracy. Furthermore, we also provide long-term results on several representative datasets with $P \in \{96, 192, 336, 720\}$, which are detailed in Table \ref{tab:long_forecasting_partial}.

\begin{table*}[htbp]
\caption{Comprehensive evaluation of multivariate forecasting performance across diverse real-world domains. \textbf{\textit{AVG}} refers to the overall average across all horizons. The \best{best} and \second{second-best} performances are highlighted. Full results can be found in Table \ref{tab:full_long_term_forecasting_results}.}
\label{tab:summary_forecasting_results}
  \vskip 0.05in
  \centering
  \resizebox{\textwidth}{!}{ 
  \begin{threeparttable}
  \begin{small}
  \renewcommand{\multirowsetup}{\centering}
  \renewcommand{\arraystretch}{1.0} 
  \setlength{\tabcolsep}{1.5pt} 
  \begin{tabular}{c|c|cc|cc|cc|cc|cc|cc|cc|cc|cc}
    \toprule
    \multicolumn{2}{c}{\multirow{2}{*}{\textbf{Models}}} & \multicolumn{2}{c}{\textbf{\method}} &
    \multicolumn{2}{c}{\textbf{Amplifier}} & \multicolumn{2}{c}{\textbf{AMD}} & \multicolumn{2}{c}{\textbf{TimeMixer}} &
    \multicolumn{2}{c}{\textbf{PatchTST}} & \multicolumn{2}{c}{\scalebox{0.9}{\textbf{iTransformer}}} & \multicolumn{2}{c}{\textbf{TSMixer}} & 
    \multicolumn{2}{c}{\textbf{DLinear}} & \multicolumn{2}{c}{\textbf{TimesNet}} \\
    \multicolumn{2}{c}{} & \multicolumn{2}{c}{\scalebox{0.8}{\textbf{(Ours)}}} & 
    \multicolumn{2}{c}{\scalebox{0.8}{(\citeyear{fei2025amplifier})}} & \multicolumn{2}{c}{\scalebox{0.8}{(\citeyear{hu2025adaptive})}} &
    \multicolumn{2}{c}{\scalebox{0.8}{(\citeyear{wang2024timemixer})}} & \multicolumn{2}{c}{\scalebox{0.8}{(\citeyear{Yuqietal-2023-PatchTST})}}&
    \multicolumn{2}{c}{\scalebox{0.8}{(\citeyear{liu2024itransformer})}}& \multicolumn{2}{c}{\scalebox{0.8}{(\citeyear{chen2023tsmixer})}}&
    \multicolumn{2}{c}{\scalebox{0.8}{(\citeyear{zeng2023transformers})}}& \multicolumn{2}{c}{\scalebox{0.8}{(\citeyear{wu2023timesnet})}} \\
    \cmidrule(lr){1-2} \cmidrule(lr){3-4} \cmidrule(lr){5-6}\cmidrule(lr){7-8} \cmidrule(lr){9-10}\cmidrule(lr){11-12}\cmidrule(lr){13-14}\cmidrule(lr){15-16}\cmidrule(lr){17-18}\cmidrule(lr){19-20}
    
    \textbf{Domain} & \textbf{Dataset} & \textbf{MSE} & \textbf{MAE} & \textbf{MSE} & \textbf{MAE} & \textbf{MSE} & \textbf{MAE} & \textbf{MSE} & \textbf{MAE} & \textbf{MSE} & \textbf{MAE} & \makebox[2.5em]{\textbf{MSE}} & \makebox[2.5em]{\textbf{MAE}} & \textbf{MSE} & \textbf{MAE} & \textbf{MSE} & \textbf{MAE} & \textbf{MSE} & \textbf{MAE} \\
    \midrule
    
    \textit{\textbf{Weather}} & \textit{Weather} & \best{0.113} & \best{0.143} & \second{0.114} & \second{0.151} & 0.129 & 0.173 & 0.115 & \second{0.151} & 0.125 & 0.164 & 0.121 & 0.155 & 0.130 & 0.195 & 0.140 & 0.197 & 0.125 & 0.175 \\
    \cmidrule{1-20}
    
    \multirow{5}{*}{\textit{\textbf{Energy}}} 
    & \textit{Electricity} & \best{0.117} & \best{0.213} & \second{0.122} & 0.219 & 0.145 & 0.240 & 0.131 & 0.226 & 0.158 & 0.253 & \second{0.122} & \second{0.215} & 0.165 & 0.276 & 0.196 & 0.288 & 0.143 & 0.250 \\
    & \textit{ETTh1} & \best{0.316} & \best{0.354} & 0.322 & 0.362 & 0.322 & \second{0.360} & 0.323 & 0.365 & \second{0.321} & 0.366 & 0.333 & 0.373 & 0.397 & 0.437 & 0.346 & 0.382 & 0.355 & 0.394 \\
    & \textit{ETTh2} & \best{0.197} & \best{0.275} & 0.206 & 0.286 & \best{0.197} & \second{0.280} & \second{0.201} & 0.283 & 0.209 & 0.291 & 0.213 & 0.294 & 0.398 & 0.475 & 0.227 & 0.317 & 0.224 & 0.305 \\
    & \textit{ETTm1} & \best{0.244} & \best{0.304} & \second{0.247} & 0.311 & 0.258 & 0.319 & 0.248 & \second{0.310} & 0.251 & 0.315 & 0.272 & 0.329 & 0.313 & 0.365 & 0.281 & 0.332 & 0.270 & 0.328 \\
    & \textit{ETTm2} & \best{0.118} & \best{0.209} & \second{0.120} & \second{0.212} & 0.127 & 0.226 & \best{0.118} & \second{0.212} & \second{0.120} & 0.214 & 0.123 & 0.217 & 0.162 & 0.285 & 0.129 & 0.235 & 0.126 & 0.220 \\
    \cmidrule{1-20}

    \multirow{2}{*}{\textit{\textbf{Traffic}}} 
    & \textit{PEMS04} & 0.129 & \best{0.229} & 0.187 & 0.298 & 0.237 & 0.333 & 0.138 & 0.249 & 0.227 & 0.331 & 0.140 & 0.246 & \best{0.114} & \second{0.237} & 0.250 & 0.358 & \second{0.117} & \best{0.229} \\
    & \textit{PEMS07} & 0.114 & \best{0.207} & 0.178 & 0.287 & 0.202 & 0.309 & 0.127 & 0.235 & 0.187 & 0.294 & \second{0.109} & \second{0.213} & \best{0.103} & 0.221 & 0.260 & 0.360 & \second{0.109} & 0.214 \\
    \cmidrule{1-20}

    \multirow{2}{*}{\textit{\textbf{Health}}} 
    & \textit{ILI} & \best{1.556} & \best{0.738} & \second{1.686} & \second{0.806} & 2.248 & 0.937 & 2.121 & 0.883 & 2.166 & 0.878 & 1.791 & 0.855 & 2.845 & 1.145 & 4.735 & 1.633 & 1.814 & 0.850 \\
    & \textit{COVID} & \second{4.071} & \best{1.174} & 4.302 & 1.263 & 5.010 & 1.304 & 10.017 & 2.012 & 4.317 & 1.243 & \best{4.054} & \second{1.230} & 8.932 & 1.860 & 10.160 & 2.001 & 9.181 & 1.872 \\
    \cmidrule{1-20}

    \textit{\textbf{Web}} & \textit{Wiki} & \second{6.702} & \best{0.440} & 6.854 & 0.505 & 6.886 & 0.472 & 6.912 & 0.510 & 6.727 & 0.461 & \best{6.687} & \second{0.448} & 7.892 & 0.838 & 6.905 & 0.547 & 7.176 & 0.569 \\
    \cmidrule{1-20}

    \multirow{2}{*}{\textit{\textbf{Finance}}} 
    & \textit{Exchange} & \best{0.035} & \best{0.125} & 0.039 & 0.132 & \second{0.036} & 0.128 & 0.037 & 0.129 & \second{0.036} & \second{0.126} & 0.040 & 0.136 & 0.105 & 0.248 & 0.044 & 0.152 & 0.047 & 0.153 \\
    & \textit{SP500} & \best{0.158} & \best{0.278} & 0.182 & 0.306 & 0.181 & 0.307 & \second{0.172} & 0.299 & 0.189 & 0.319 & 0.224 & 0.352 & 0.263 & 0.387 & 0.268 & 0.407 & 0.179 & \second{0.298} \\
    
    \midrule
    \rowcolor{rowblue} \multicolumn{2}{c|}{\textbf{\textit{AVG}}} & \best{1.067} & \best{0.361} & 1.120 & 0.395 & 1.229 & 0.414 & 1.589 & 0.451 & 1.156 & 0.404 & \second{1.095} & \second{0.390} & 1.678 & 0.536 & 1.842 & 0.554 & 1.528 & 0.450 \\
    \midrule
    \rowcolor{rowpink} \multicolumn{2}{c|}{\textit{\textbf{1\textsuperscript{st} Count}}} 
    & \multicolumn{2}{c|}{\best{22}} & \multicolumn{2}{c|}{0} & \multicolumn{2}{c|}{1} & \multicolumn{2}{c|}{1} & \multicolumn{2}{c|}{0} & \multicolumn{2}{c|}{\second{2}} & \multicolumn{2}{c|}{\second{2}} & \multicolumn{2}{c|}{0} & \multicolumn{2}{c}{1} \\
    \bottomrule
  \end{tabular}
  \end{small}
  \end{threeparttable}
  }
\end{table*}

\noindent\textbf{Results.} Table~\ref{tab:summary_forecasting_results} shows \method\ outperforms other models across most real-world multivariate domains. On average, it surpasses the second-best model by \textbf{2.6\%} in MSE and \textbf{7.4\%} in MAE. Specifically, on the high-dimensional Electricity dataset, \method\ achieves a \textbf{4.1\%} MSE reduction over the strongest baseline. For challenging datasets with severe fluctuations, it exceeds the second-best alternatives by \textbf{7.7\%} in MSE on Healthcare (ILI) and \textbf{8.1\%} in MSE on Finance (SP500).

\subsubsection{Univariate Forecasting}
\label{subsec:real_uni}

\noindent\textbf{Setups.} To conduct a comprehensive evaluation, we assess \method\ on the \textbf{M4 Competition dataset} \citep{makridakis2018m4}. This dataset consists of 100,000 real-world time series covering multiple temporal frequencies, ranging from hourly to yearly. Its large scale and heterogeneous characteristics provide a challenging benchmark for evaluating the robustness and generalization ability of forecasting models.

\begin{table}[htbp!]
\caption{Evaluation of univariate forecasting performance on the M4 Competition dataset. 
The \best{best} and \second{second-best} performances are highlighted. 
Full results can be found in Table \ref{tab:m4_results_univariate}.}
\label{tab:m4_main_results}
\vskip 0.05in
\centering
\resizebox{\columnwidth}{!}{
\begin{threeparttable}
\normalsize
\renewcommand{\arraystretch}{1.0}
\setlength{\tabcolsep}{1.0pt}

\newcommand{\num}[1]{{\large #1}}
\newcommand{\met}[1]{{\textbf{#1}}}
\newcommand{\modelhead}[1]{\scalebox{1.00}{#1}}
\newcommand{\yearhead}[1]{\scalebox{0.95}{#1}}

\begin{tabular}{c|cccccccccccccccc}
\toprule
\multicolumn{1}{c}{\multirow{2}{*}{Models}} &
\multicolumn{1}{c}{\modelhead{\textbf{\method}}} &
\multicolumn{1}{c}{\modelhead{\textbf{Amplifier}}} &
\multicolumn{1}{c}{\modelhead{\textbf{TimeMixer}}} &
\multicolumn{1}{c}{\modelhead{\textbf{iTrans.}}} &
\multicolumn{1}{c}{\modelhead{\textbf{TiDE}}} &
\multicolumn{1}{c}{\modelhead{\textbf{TNet}}} &
\multicolumn{1}{c}{\modelhead{\textbf{N-HiTS}}} &
\multicolumn{1}{c}{\modelhead{\textbf{DLinear}}} &
\multicolumn{1}{c}{\modelhead{\textbf{Patch}}} &
\multicolumn{1}{c}{\modelhead{\textbf{MICN}}} &
\multicolumn{1}{c}{\modelhead{\textbf{FiLM}}} &
\multicolumn{1}{c}{\modelhead{\textbf{LightTS}}} &
\multicolumn{1}{c}{\modelhead{\textbf{FED}}} &
\multicolumn{1}{c}{\modelhead{\textbf{Stat.}}} &
\multicolumn{1}{c}{\modelhead{\textbf{Auto}}} &
\multicolumn{1}{c}{\modelhead{\textbf{N-BEATS}}} \\

\multicolumn{1}{c}{} &
\multicolumn{1}{c}{\yearhead{(Ours)}} &
\multicolumn{1}{c}{\yearhead{(\citeyear{fei2025amplifier})}} &
\multicolumn{1}{c}{\yearhead{(\citeyear{wang2024timemixer})}} &
\multicolumn{1}{c}{\yearhead{(\citeyear{liu2024itransformer})}} &
\multicolumn{1}{c}{\yearhead{(\citeyear{das2023longterm})}} &
\multicolumn{1}{c}{\yearhead{(\citeyear{wu2023timesnet})}} &
\multicolumn{1}{c}{\yearhead{(\citeyear{challu2023nhits})}} &
\multicolumn{1}{c}{\yearhead{(\citeyear{zeng2023transformers})}} &
\multicolumn{1}{c}{\yearhead{(\citeyear{Yuqietal-2023-PatchTST})}} &
\multicolumn{1}{c}{\yearhead{(\citeyear{wang2023micn})}} &
\multicolumn{1}{c}{\yearhead{(\citeyear{zhou2022film})}} &
\multicolumn{1}{c}{\yearhead{(\citeyear{campos2023lightts})}} &
\multicolumn{1}{c}{\yearhead{(\citeyear{zhou2022fedformer})}} &
\multicolumn{1}{c}{\yearhead{(\citeyear{liu2022non})}} &
\multicolumn{1}{c}{\yearhead{(\citeyear{wu2021autoformer})}} &
\multicolumn{1}{c}{\yearhead{(\citeyear{Oreshkin2020N-BEATS})}} \\
\toprule

\met{SMAPE}
& \best{\num{11.671}} & \num{12.030} & \second{\num{11.723}} & \num{12.684} & \num{13.950}
& \num{11.829} & \num{11.927} & \num{13.639} & \num{13.152} & \num{19.638}
& \num{14.863} & \num{13.525} & \num{12.840} & \num{12.780} & \num{12.909} & \num{11.851} \\

\met{MASE}
& \best{\num{1.546}} & \num{1.617} & \second{\num{1.559}} & \num{1.764} & \num{1.940}
& \num{1.585} & \num{1.613} & \num{2.095} & \num{1.945} & \num{5.947}
& \num{2.207} & \num{2.111} & \num{1.701} & \num{1.756} & \num{1.771} & \second{\num{1.559}} \\

\rowcolor{rowblue}
\met{OWA}
& \best{\num{0.834}} & \num{0.887} & \second{\num{0.840}} & \num{0.929} & \num{1.020}
& \num{0.851} & \num{0.861} & \num{1.051} & \num{0.998} & \num{2.279}
& \num{1.125} & \num{1.051} & \num{0.918} & \num{0.930} & \num{0.939} & \num{0.855} \\
\bottomrule
\end{tabular}
\end{threeparttable}
}
\end{table}

\noindent\textbf{Results.} As shown in Table~\ref{tab:m4_main_results}, \method\ achieves the best performance across all three metrics on the M4 dataset. 
It obtains the lowest SMAPE (11.671), MASE (1.546), and OWA (0.834), outperforming all compared baselines. 
In particular, \method\ achieves the best OWA, which is the primary metric of the M4 competition. This result confirms strong overall forecasting capability.

\subsection{Evaluation on Synthetic Datasets}
\label{sec:sim_forecasting}

\begin{wrapfigure}{r}{0.65\columnwidth}
\vspace{-4mm}
\centering
\includegraphics[width=1\linewidth]{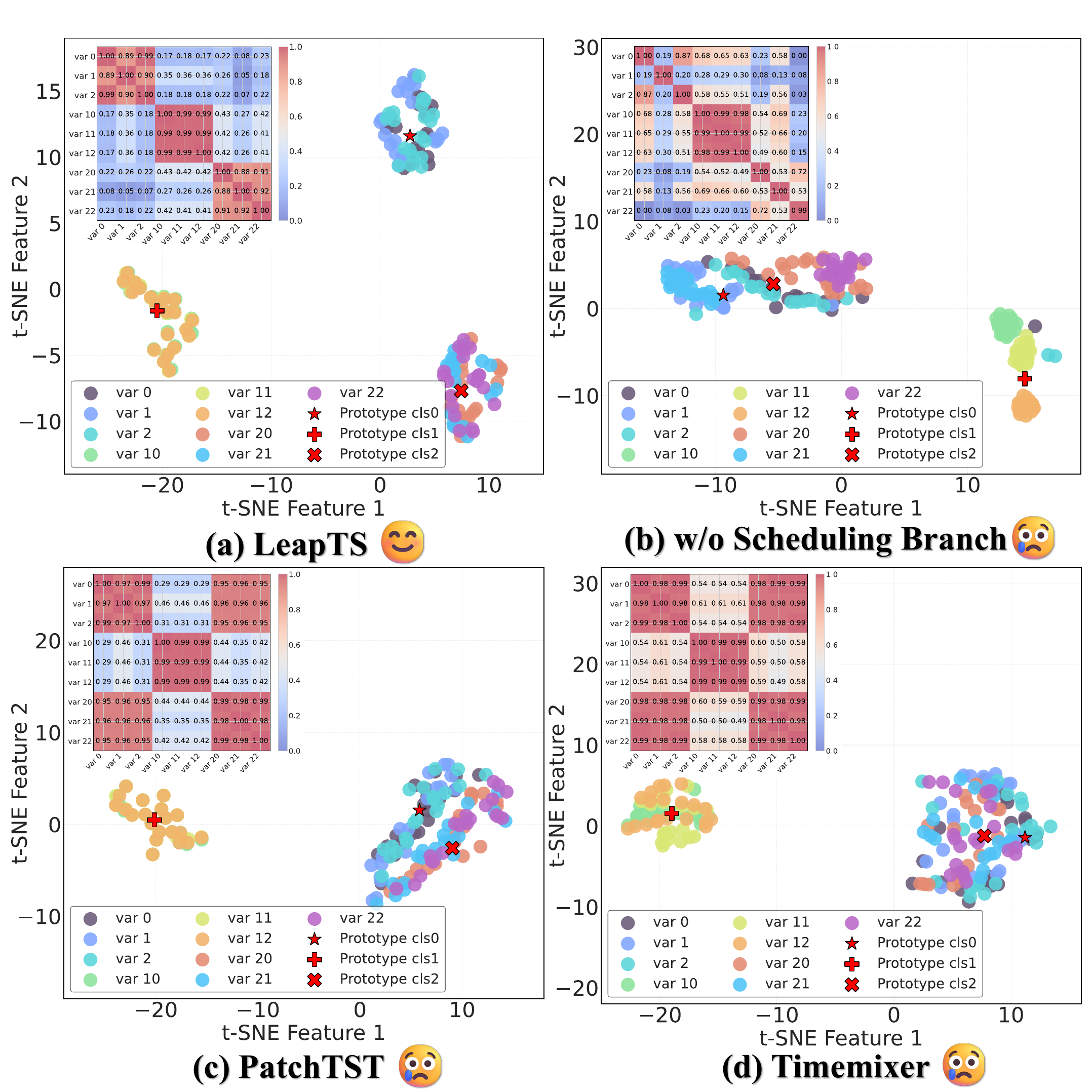}
\vspace{-1mm}
\caption{t-SNE visualization in Scenario 1. \method\ successfully distinguishes the variable clusters, whereas baselines fail.}
\label{fig:tsne}
\vspace{-4mm}
\end{wrapfigure}

\paragraph{Setups.} We construct three synthetic datasets to simulate specific temporal dynamics (details in Appendix \ref{appendx:Syntheti}). \textbf{Scenario 1} simulates structurally clustered \textbf{high-dimensional correlated variables}. \textbf{Scenario 2} focuses on multivariate sequences driven by \textbf{explicit causal interactions}. \textbf{Scenario 3} introduces univariate series influenced by \textbf{unobserved factors}. These three distinct settings comprehensively assess the model's forecasting robustness across highly complex environments.

\noindent\textbf{Results.} Fig.~\ref{fig:tsne} shows the t-SNE embeddings of hidden representations from three distinct variable clusters. The plot indicates that only \method\ \textbf{\textit{successfully separates}} these groups in the latent space. In contrast, the compared baselines \textbf{\textit{fail to distinguish}} these features and tend to mix variables from cluster~0 and cluster~1. Table~\ref{tab:synthetic_full_results} shows that \method\ outperforms the advanced baselines by large margins. Specifically, it reduces the average MSE by \textbf{50.0\%} (0.088$\rightarrow$0.044) on Scenario~1, \textbf{80.5\%} (0.329$\rightarrow$0.064) on Scenario~2, and \textbf{90.7\%} (0.193$\rightarrow$0.018) on Scenario~3. These substantial improvements demonstrate the strong robustness of \method\ across diverse complex environments. Furthermore, the sharp performance drop in the \textit{w/o scheduling branch} variant indicates the importance of the proposed scheduling mechanism.

\begin{table*}[htbp]
    \caption{Full forecasting results on three synthetic datasets: (1) high-dimensional heterogeneous series, (2) causally driven series, and (3) univariate series with latent dynamics. Baselines are optimized over an wider hyperparameter search space (see Appendix \ref{appendx:imp_detail}). \textbf{\textit{AVG}} refers to the average across all prediction lengths. The \best{best} and \second{second-best} performances are highlighted.}
    \label{tab:synthetic_full_results}
  \vskip 0.05in
  \centering
  \resizebox{\textwidth}{!}{
  \begin{threeparttable}
  \begin{small}
  \renewcommand{\arraystretch}{1}
  \setlength{\tabcolsep}{3.5pt}
  \begin{tabular}{c|c|ccccc|ccccc|ccccc}
    \toprule
    \multirow{2}{*}{\textbf{Models}} & \multirow{2}{*}{\textbf{Metric}}
    & \multicolumn{5}{c|}{\cellcolor{black!10}\textbf{Scenario 1 (Multivariate)}}
    & \multicolumn{5}{c|}{\cellcolor{black!10}\textbf{Scenario 2 (Multivariate)}}
    & \multicolumn{5}{c}{\cellcolor{black!10}\textbf{Scenario 3 (Univariate)}} \\
    & & \textbf{24} & \textbf{36} & \textbf{48} & \textbf{60} & \textbf{\textit{AVG}}
      & \textbf{24} & \textbf{36} & \textbf{48} & \textbf{60} & \textbf{\textit{AVG}}
      & \textbf{24} & \textbf{36} & \textbf{48} & \textbf{60} & \textbf{\textit{AVG}} \\
    \midrule

    \multirow{2}{*}{\makecell{\textbf{\method} \\ \textbf{\textit{(Ours)}}}}
    & \textbf{MSE}
    & \best{0.027} & \best{0.040} & \best{0.051} & \best{0.058} & \cellcolor{rowblue}\best{0.044}
    & \best{0.033} & \best{0.050} & \best{0.074} & \best{0.099} & \cellcolor{rowblue}\best{0.064}
    & \best{0.012} & \best{0.016} & \best{0.021} & \best{0.024} & \cellcolor{rowblue}\best{0.018} \\
    & \textbf{MAE}
    & \best{0.022} & \best{0.028} & \best{0.034} & \best{0.038} & \cellcolor{rowblue}\best{0.030}
    & \best{0.059} & \best{0.083} & \best{0.102} & \best{0.126} & \cellcolor{rowblue}\best{0.092}
    & \best{0.058} & \best{0.070} & \best{0.080} & \best{0.090} & \cellcolor{rowblue}\best{0.074} \\
    \midrule

    \multirow{2}{*}{\makecell{\textit{\textbf{w/o Scheduling}} \\ \textbf{Branch}}}
    & \textbf{MSE}
    & 0.145 & 0.277 & 0.421 & 0.509 & \cellcolor{rowblue}0.338
    & 0.390 & 0.526 & 0.623 & 0.692 & \cellcolor{rowblue}0.558
    & 0.443 & 0.511 & 0.551 & 0.582 & \cellcolor{rowblue}0.522 \\
    & \textbf{MAE}
    & 0.111 & 0.174 & 0.257 & 0.304 & \cellcolor{rowblue}0.212
    & 0.257 & 0.335 & 0.395 & 0.443 & \cellcolor{rowblue}0.358
    & 0.440 & 0.491 & 0.519 & 0.541 & \cellcolor{rowblue}0.498 \\
    \midrule

    \multirow{2}{*}{\textit{\textbf{PatchTST}}}
    & \textbf{MSE}
    & \second{0.059} & \second{0.082} & \second{0.097} & \second{0.114} & \cellcolor{rowblue}\second{0.088}
    & \second{0.192} & \second{0.294} & \second{0.391} & \second{0.440} & \cellcolor{rowblue}\second{0.329}
    & \second{0.116} & \second{0.173} & \second{0.217} & \second{0.266} & \cellcolor{rowblue}\second{0.193} \\
    & \textbf{MAE}
    & \second{0.093} & \second{0.100} & \second{0.109} & \second{0.113} & \cellcolor{rowblue}\second{0.104}
    & \second{0.192} & \second{0.253} & \second{0.299} & \second{0.321} & \cellcolor{rowblue}\second{0.266}
    & \second{0.185} & \second{0.240} & \second{0.277} & \second{0.314} & \cellcolor{rowblue}\second{0.254} \\
    \midrule

    \multirow{2}{*}{\textit{\textbf{TimeMixer}}}
    & \textbf{MSE}
    & 0.104 & 0.197 & 0.282 & 0.371 & \cellcolor{rowblue}0.239
    & 0.375 & 0.496 & 0.579 & 0.640 & \cellcolor{rowblue}0.523
    & 0.251 & 0.335 & 0.418 & 0.456 & \cellcolor{rowblue}0.365 \\
    & \textbf{MAE}
    & 0.117 & 0.176 & 0.226 & 0.276 & \cellcolor{rowblue}0.199
    & 0.225 & 0.289 & 0.348 & 0.387 & \cellcolor{rowblue}0.312
    & 0.316 & 0.388 & 0.449 & 0.476 & \cellcolor{rowblue}0.407 \\

    \bottomrule
  \end{tabular}
  \end{small}
  \end{threeparttable}
  }
\end{table*}

\begin{wrapfigure}{r}{0.6\columnwidth}
\vspace{-4mm}
\centering
\includegraphics[width=\linewidth]{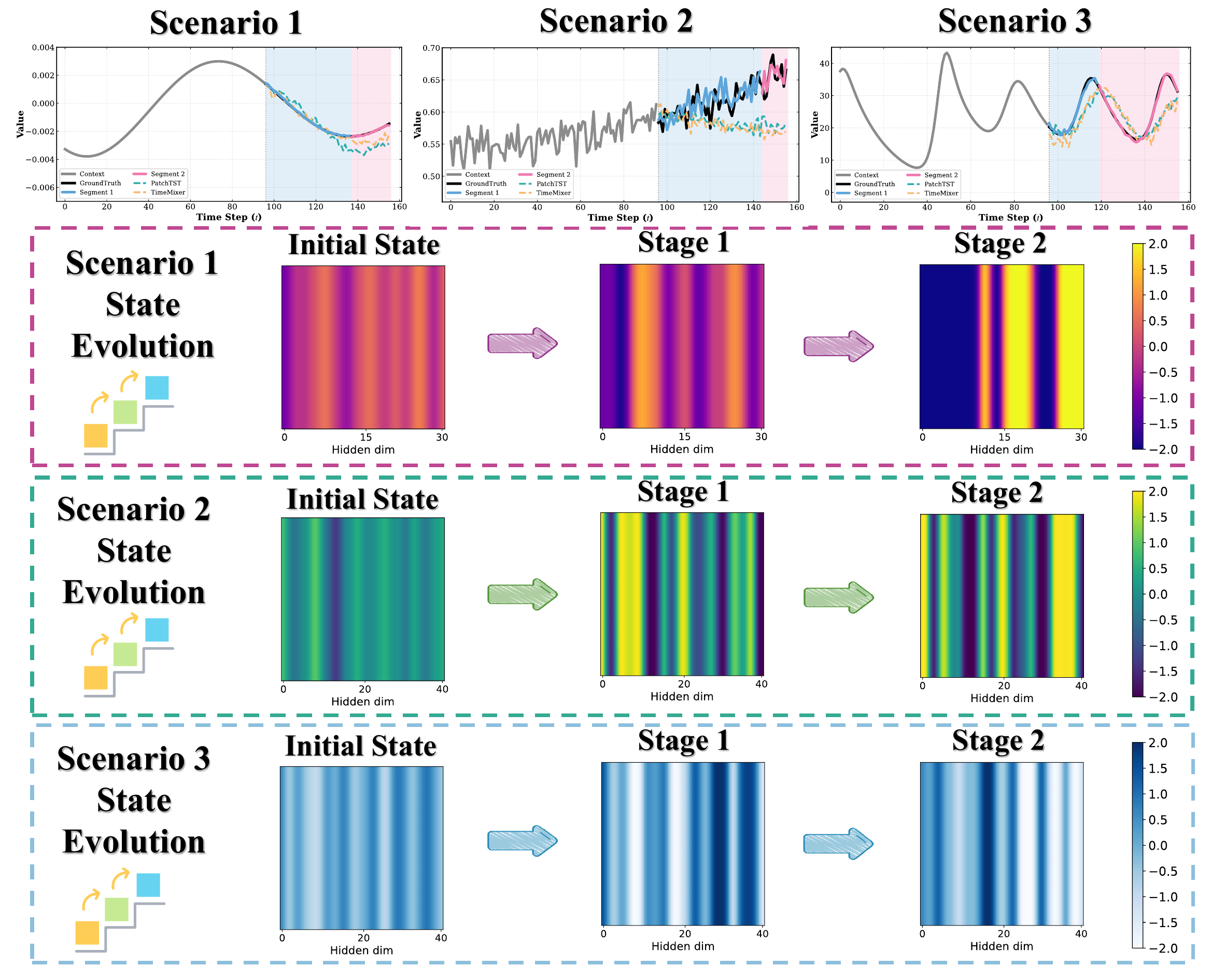}
\vspace{-1mm}
\caption{Visualization of forecasting behavior and latent-state evolution across three synthetic scenarios.}
\vspace{-20mm}
\label{fig:synthetic_vis}
\end{wrapfigure}

Fig.~\ref{fig:synthetic_vis} and Fig.~\ref{decision_fig3} provide qualitative evidence supporting these results. Around the \textbf{\textit{turning points}} in Scenario~1 and Scenario~3, \method\ closely follows the real temporal pattern, while competing models often \textit{lag behind these changes}. In Scenario~2, both baselines simply reproduce the \textit{historical mean-level fluctuations}, whereas \method\ successfully captures the \textbf{\textit{emerging upward trend}}. The visualization also reveals that \method\ exhibits an \textbf{\textit{adaptive scheduling behavior}}. As forecasting progresses, the model dynamically leverages information from different context segments.

\subsection{Scheduling Analysis}
\label{sec:explainability}

\begin{wraptable}{r}{0.6\textwidth}
\centering
\caption{Proportions of control (\textbf{Ct.}) and temporal (\textbf{Te.}) signals across four bins of increasing volatility, ranging from Bin 0 (lowest) to Bin 3 (highest). The dominant \textbf{Overall} per dataset is in \underline{\textcolor{secondcolor}{blue}}, and the highest \textbf{\textit{AVG}} is in \textbf{\textcolor{bestcolor}{orange}}.}
\label{tab:bin_analysis}
\begin{threeparttable}
\resizebox{0.6\textwidth}{!}{
\renewcommand{\arraystretch}{0.95}
\setlength{\tabcolsep}{3pt}
\begin{tabular}{l|cc|cc|cc|cc|cc}
\toprule
\multirow{2}{*}{\textbf{Dataset}} 
& \multicolumn{2}{c|}{\textbf{Bin 0}} 
& \multicolumn{2}{c|}{\textbf{Bin 1}} 
& \multicolumn{2}{c|}{\textbf{Bin 2}} 
& \multicolumn{2}{c|}{\textbf{Bin 3}} 
& \multicolumn{2}{c}{\textbf{Overall}} \\
\cmidrule(lr){2-3} \cmidrule(lr){4-5} \cmidrule(lr){6-7} \cmidrule(lr){8-9} \cmidrule(lr){10-11}
& \textbf{Ct.} & \textbf{Te.} & \textbf{Ct.} & \textbf{Te.} & \textbf{Ct.} & \textbf{Te.} & \textbf{Ct.} & \textbf{Te.} & \textbf{Ct.} & \textbf{Te.} \\
\midrule
\textbf{\textit{Elec.}} & 0.216 & 0.784 & 0.269 & 0.731 & 0.277 & 0.723 & 0.272 & 0.728 & \cellcolor{rowpink}0.259 & \cellcolor{rowpink}\second{0.741} \\
\textbf{\textit{ETTh2}} & 0.530 & 0.471 & 0.572 & 0.428 & 0.585 & 0.415 & 0.604 & 0.396 & \cellcolor{rowpink}\second{0.573} & \cellcolor{rowpink}0.427 \\
\textbf{\textit{ETTm2}} & 0.359 & 0.641 & 0.431 & 0.569 & 0.467 & 0.533 & 0.473 & 0.527 & \cellcolor{rowpink}0.432 & \cellcolor{rowpink}\second{0.568} \\
\textbf{\textit{Exch.}} & 0.947 & 0.053 & 0.945 & 0.055 & 0.943 & 0.057 & 0.942 & 0.058 & \cellcolor{rowpink}\second{0.944} & \cellcolor{rowpink}0.056 \\
\textbf{\textit{PEMS.}} & 0.260 & 0.740 & 0.241 & 0.759 & 0.234 & 0.766 & 0.265 & 0.735 & \cellcolor{rowpink}0.250 & \cellcolor{rowpink}\second{0.750} \\
\midrule
\rowcolor{rowblue}
\textbf{\textit{AVG}} & \textbf{0.462} & \textbf{\textcolor{bestcolor}{0.538}} & \textbf{0.492} & \textbf{0.508} & \textbf{0.501} & \textbf{0.499} & \textbf{\textcolor{bestcolor}{0.511}} & \textbf{0.489} & \textbf{0.492} & \textbf{0.508} \\
\bottomrule
\end{tabular}
}
\begin{tablenotes}
    \tiny
    \item Note: \textbf{Ct.}: Control signal; \textbf{Te.}: Temporal signal.
\end{tablenotes}
\end{threeparttable}
\vspace{-2mm}
\end{wraptable}

\paragraph{Scheduling Signal Decomposition.} To analyze the internal \textbf{scheduling mechanism}, we decompose the total signal into \textbf{control} and \textbf{temporal} components (see Appendix~\ref{app:decomposition} for specific details). The relative proportions of these signals are then calculated to identify the dominant driver. Specifically, we partition all windows into four bins in ascending order of volatility. Table~\ref{tab:bin_analysis} illustrates the shift in signal proportions across different volatility regimes.  At the \textbf{window level}, temporal signals dominate low-volatility inputs (\textbf{0.538}, Bin 0), while control signals increase in high-volatility ones (\textbf{0.511}, Bin 3).  At the \textbf{dataset level}, temporal signals are more prominent in periodic data (\textit{PEMS04}: \textbf{0.750}), whereas control signals dominate more volatile data (\textit{Exchange}: \textbf{0.944}).  These observations lead to the following insight:

\begin{tcolorbox}[boxsep=0mm,left=1.2mm,right=1.2mm,colframe=black!55,colback=black!5]
{\textit{\textbf{Insight 1:}}
The dominant signal varies with volatility levels and dataset characteristics: temporal signals dominate stable or periodic regimes, while control signals dominate volatile regimes.} 
\end{tcolorbox}

\paragraph{Volatility-Aware Scheduling.} 

\begin{wrapfigure}{r}{0.6\columnwidth}
\vspace{-8mm}
\centering
\includegraphics[width=1.0\linewidth]{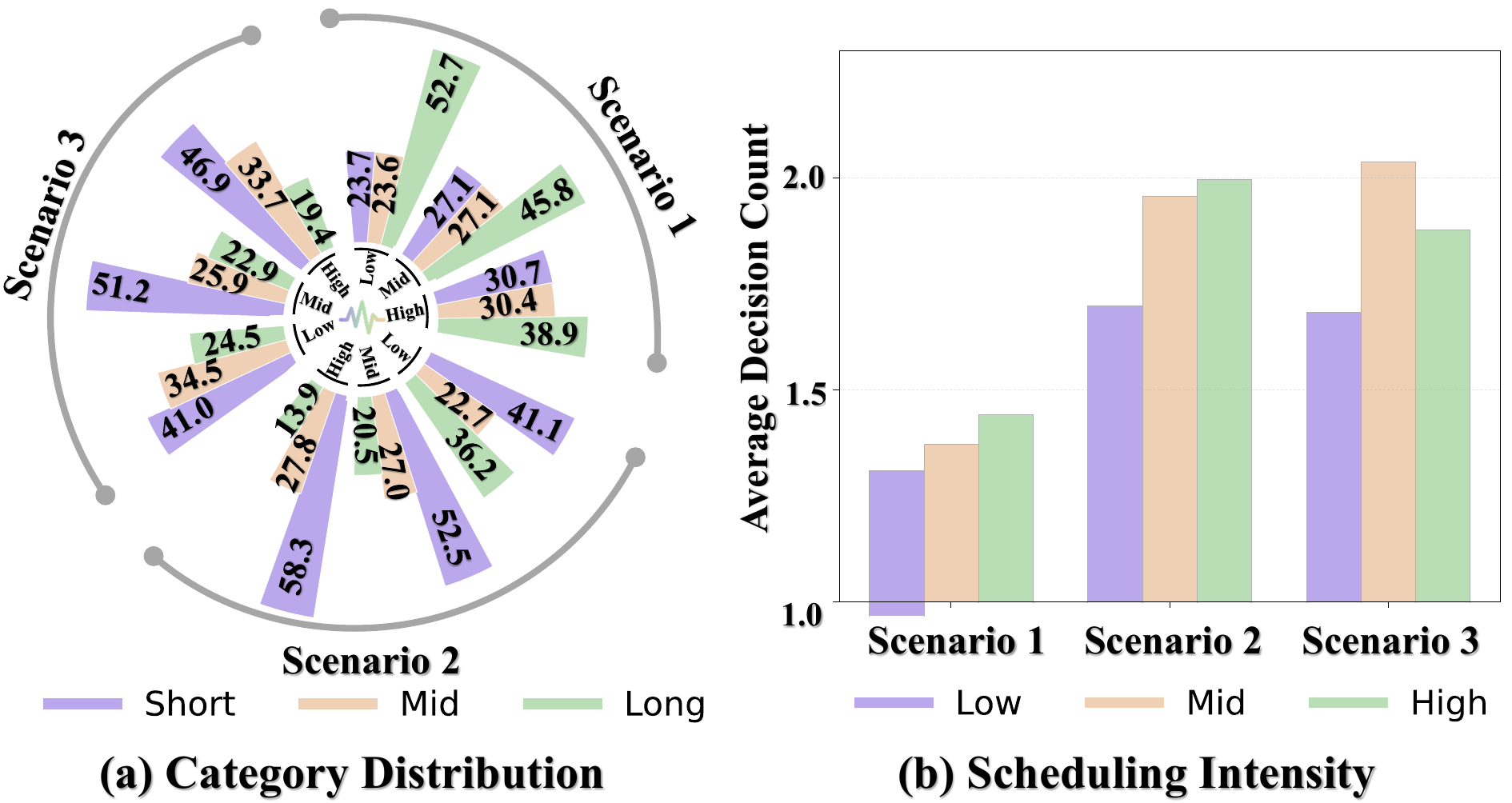}
\vspace{-2.5mm}
\caption{Scheduling behaviors on synthetic data. \textbf{(a) Category Distribution:} proportions across volatility levels. \textbf{(b) Scheduling Intensity:} average scheduling count.}\label{fig:synthetic}
\vspace{-3mm}
\end{wrapfigure}

We analyze scheduling patterns under three synthetic scenarios. As volatility increases, the \textbf{category distribution} (Fig.~\ref{fig:synthetic}a) shifts toward \textbf{Short} scheduling decisions.  For example, in Scenario 2 the proportion rises from \textbf{41.1\%} to \textbf{58.3\%}, and similar trends appear in the other scenarios.  This indicates that the model becomes more \textbf{cautious} under higher uncertainty.  Meanwhile, the \textbf{scheduling intensity} (Fig.~\ref{fig:synthetic}b) also increases across all scenarios. In Scenario 1, the average number of scheduling decisions per window rises from \textbf{1.32} to \textbf{1.45}, indicating more frequent scheduling under higher volatility. These observations lead to the following insight:

\begin{tcolorbox}[boxsep=0mm,left=1.2mm,right=1.2mm,colframe=black!55,colback=black!5]
{\textit{\textbf{Insight 2:}}
Scheduling adapts to volatility: low-volatility regimes tend toward longer categories and fewer actions, while high-volatility regimes toward shorter categories and more frequent actions.} 
\end{tcolorbox}

\subsection{Additional Analysis}
\label{sec:ad_ana}

\paragraph{Efficiency Analysis}
\begin{wrapfigure}{r}{0.7\columnwidth}
\vspace{-4mm}
\centering
\includegraphics[width=1.0\linewidth]{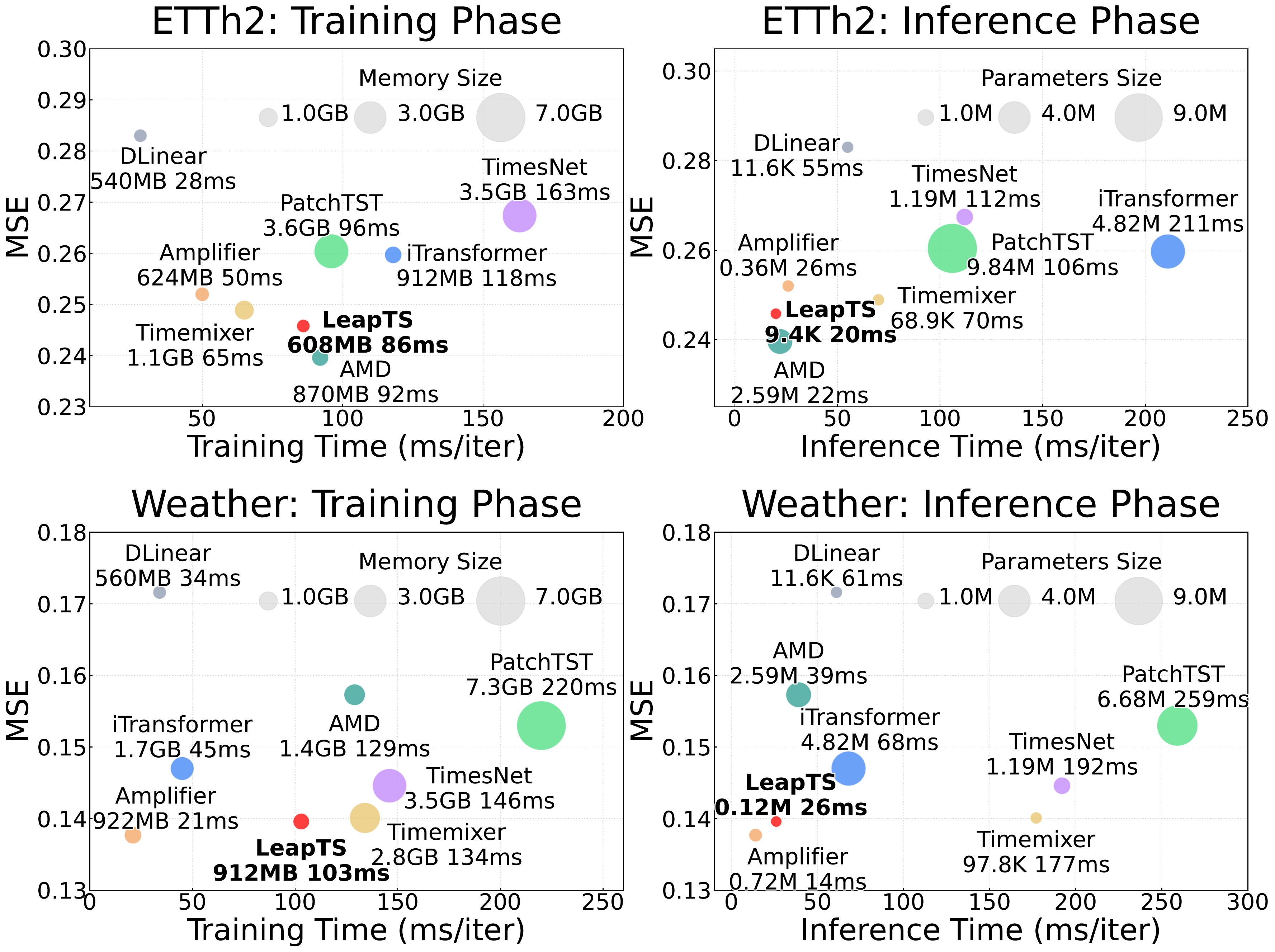}
\vspace{-2.5mm}
\caption{Training and inference efficiency on ETTh2 and Weather. Bubble size indicates memory (training) or parameter size (inference).}\label{fig:efficiency}
\vspace{-5mm}
\end{wrapfigure}
Fig.~\ref{fig:efficiency} compares training and inference efficiency on ETTh2 and Weather. Our method achieves lower MSE while maintaining competitive training and inference time. It also requires less memory during training and fewer parameters during inference, demonstrating strong efficiency alongside improved accuracy.

\noindent\textbf{Ablation Study} Table~\ref{tab:ablation_study} summarizes the effects of key components. Removing the scheduling branch causes the largest performance drop, with average MSE increasing by \textbf{22.22\%}, which highlights the importance of adaptive scheduling. Replacing the CDE with RNN or LSTM also leads to consistent degradation. This suggests that the continuous-time formulation better captures irregular time evolution. 

\begin{wraptable}{r}{0.35\textwidth}
\vspace{-4mm} 
\caption{Ablation study on the scheduling trace.}
\label{tab:ablation_trace_main}
\centering
\resizebox{\linewidth}{!}{
\begin{tabular}{lcc}
\toprule
\textbf{Method} & \textbf{Average} & \textit{\textbf{Improv.}} \\
\midrule
\rowcolor{rowblue}
\method & \textbf{0.323} & - \\
Monte Carlo & 0.378 & \textcolor{bestcolor}{\textbf{\textit{+14.55\%}}} \\
Fixed Step & 0.370 & \textcolor{bestcolor}{\textbf{\textit{+12.70\%}}} \\
\bottomrule
\end{tabular}
}
\vspace{-2mm} 
\end{wraptable}
Furthermore, we evaluate the scheduling trace against Monte Carlo and Fixed Step schemes. Table~\ref{tab:ablation_trace_main} shows that periodicity-based fixed steps yield marginal gains over Monte Carlo. In contrast, \method\ outperforms both baselines and reduces average MAE by \textbf{14.55\%} and \textbf{12.70\%}, respectively. These results verify the effectiveness of the adaptive scheduling trace. Appendix~\ref{appendx:ablation} provides experimental settings and full result descriptions for this analysis.

\begin{table}[htbp]
\caption{Ablation study on model components. We evaluate removing the scheduling branch and replacing the CDE cell with RNN/LSTM. Results report average MSE across prediction lengths.}
\label{tab:ablation_study}
\vskip 0.02in
\centering
\resizebox{\columnwidth}{!}{
\begin{tabular}{lcccccccccc}
\toprule
 & Weather & Electricity & ETTh1 & ETTh2 & ETTm1 & ETTm2 & PEMS04 & PEMS07 & Average & \textit{\textbf{Improv.}} \\
\midrule

\rowcolor{rowblue}
\method
& \textbf{0.113} 
& \textbf{0.117} 
& \textbf{0.316} 
& \textbf{0.197} 
& \textbf{0.244} 
& \textbf{0.118} 
& \textbf{0.129} 
& \textbf{0.114} 
& \textbf{0.168} 
& - \\

\hspace{2em} w/o hierarchical scheduling
& 0.140
& 0.167
& 0.330
& 0.199
& 0.278
& 0.125
& 0.247
& 0.242
& 0.216 
& \textcolor{bestcolor}{\textbf{\textit{+22.22\%}}} \\

\hspace{2em} RNN cell
& 0.114
& 0.119
& 0.320
& 0.199
& 0.248
& 0.119
& 0.136
& 0.117
& 0.172 
& \textcolor{bestcolor}{\textbf{\textit{+2.33\%}}} \\

\hspace{2em} LSTM cell
& 0.114
& 0.118
& 0.319
& 0.198
& 0.245
& 0.118
& 0.135
& 0.117
& 0.171 
& \textcolor{bestcolor}{\textbf{\textit{+1.75\%}}} \\

\bottomrule
\end{tabular}
}
\end{table}

\section{Conclusion}

In this paper, we present \method, a novel framework that reformulates time series prediction as a dynamic scheduling process. By integrating a hierarchical controller with continuous-time state evolution, \method~facilitates adaptive scheduling and effectively captures temporal dynamics. Extensive experiments on real-world and synthetic datasets demonstrate that \method~achieves superior forecasting accuracy and efficiency while exhibiting traceable scheduling behaviors. These results also suggest a promising paradigm for time series modeling.

\bibliographystyle{plainnat}
\bibliography{refs}

\newpage

\appendix

\section{Differentiability Analysis}
\label{appendx:differentiability}

\paragraph{Differentiability of the High-Level Scheduler.} 
To make the discrete category selection end-to-end trainable, we overcome the non-differentiable $\operatorname{argmax}$ operation using the Gumbel-Softmax trick with a Straight-Through (ST) estimator \citep{jang2017categorical,maddison2016concrete}. Given the unnormalized logits $\mathbf{o}_k = \mathbf{W}_{\mathrm{cat}}\mathbf{h}_k \in \mathbb{R}^C$ and independent noise $g_i \sim \operatorname{Gumbel}(0, 1)$, we first compute the distribution $\tilde{\boldsymbol{\pi}}_k$:
\begin{equation}
\tilde{\pi}_{k,i} = \frac{\exp\big((o_{k,i} + g_i) / \tau_g\big)}{\sum_{j=1}^{C} \exp\big((o_{k,j} + g_j) / \tau_g\big)},
\label{eq:gumbel_softmax}
\end{equation}
where $\tau_g > 0$ is the temperature. To ensure strict discrete routing, the forward pass uses the hard one-hot sample $\boldsymbol{\pi}_k = \operatorname{one\_hot}(\operatorname{argmax}_i \tilde{\pi}_{k,i})$. To bypass the zero gradients of $\operatorname{argmax}$ during backpropagation, the ST estimator directly routes the gradient of the loss $\mathcal{L}$ through the soft approximation:
\begin{equation}
\nabla_{\mathbf{o}_k} \mathcal{L} \approx \nabla_{\tilde{\boldsymbol{\pi}}_k} \mathcal{L} \cdot \frac{\partial \tilde{\boldsymbol{\pi}}_k}{\partial \mathbf{o}_k}.
\label{eq:st_estimator}
\end{equation}
This formulation maintains exact discrete execution forward while providing dense, continuous gradient feedback backward to optimize the decision boundary \citep{yin2018understanding}.

\paragraph{Differentiability of the Low-Level Scheduler.}
To ensure the advancement length $\ell_k$ is trainable, we replace the non-differentiable hard truncation with a continuous soft mask $m_k(\tau)$ \citep{graves2016adaptive}. For clarity, we present the derivation for one variable and omit the variable index, and the same derivation is applied independently to each variable. Let $\mathcal{L}$ denote the overall forecasting loss. The gradient with respect to $\ell_k$ is computed via the chain rule over all future steps $\tau$:
\begin{equation}
\frac{\partial \mathcal{L}}{\partial \ell_k}
=
\sum_{\tau=1}^{P}
\frac{\partial \mathcal{L}}{\partial \hat{y}^{\mathrm{sched}}_{k+1}(\tau)}
\frac{\partial \hat{y}^{\mathrm{sched}}_{k+1}(\tau)}{\partial \bar{s}_k(\tau)}
\frac{\partial \bar{s}_k(\tau)}{\partial m_k(\tau)}
\frac{\partial m_k(\tau)}{\partial \ell_k}.
\label{eq:chain_rule}
\end{equation}

Based on the iterative update $\hat{\mathbf{y}}^{\mathrm{sched}}_{k+1} = \hat{\mathbf{y}}^{\mathrm{sched}}_{k} + \bar{\mathbf{s}}_k$ and the masking operation $\bar{\mathbf{s}}_k = \mathbf{s}_k \odot \mathbf{m}_k$, the intermediate derivatives are:
\begin{equation}
\frac{\partial \hat{y}^{\mathrm{sched}}_{k+1}(\tau)}{\partial \bar{s}_k(\tau)} = 1,
\qquad
\frac{\partial \bar{s}_k(\tau)}{\partial m_k(\tau)} = s_k(\tau).
\end{equation}

Let $\sigma(\cdot)$ denote the standard logistic sigmoid function. For the soft mask $m_k(\tau) = \mathbb{I}[\tau \ge q_k] \cdot \sigma(z_{k,\tau})$, where $z_{k,\tau} = \frac{\ell_k - (\tau - q_k) - 0.5}{\gamma}$, the indicator term $\mathbb{I}[\tau \ge q_k]$ depends only on the discrete time indices. Because it acts strictly as a constant binary routing gate during backpropagation, it does not break the computational graph of $\ell_k$ \citep{maddison2016concrete}. Thus, the derivative with respect to $\ell_k$ can be derived as:
\begin{equation}
\frac{\partial m_k(\tau)}{\partial \ell_k}
=
\mathbb{I}[\tau \ge q_k]
\cdot
\frac{1}{\gamma}
\cdot
\sigma(z_{k,\tau})
\big(1-\sigma(z_{k,\tau})\big).
\end{equation}

Substituting these intermediate terms back into Eq.~\eqref{eq:chain_rule}, the indicator function restricts the summation range to $\tau \ge q_k$, formulating the final gradient flow as:
\begin{equation}
\frac{\partial \mathcal{L}}{\partial \ell_k}
=
\sum_{\tau=q_k}^{P}
\frac{\partial \mathcal{L}}
{\partial \hat{y}^{\mathrm{sched}}_{k+1}(\tau)}
s_k(\tau)
\frac{
\sigma(z_{k,\tau})
\big(1-\sigma(z_{k,\tau})\big)
}{\gamma}.
\end{equation}

This formulation provides a differentiable path from the forecasting loss to the continuous advancement length through the soft mask. For any finite $z_{k,\tau}$ and temperature $\gamma > 0$, the derivative $\sigma(z_{k,\tau})(1-\sigma(z_{k,\tau}))$ is positive, providing smooth feedback around the soft boundary and enabling optimization of the hierarchical scheduling mechanism \citep{chung2017hierarchical}.

\section{Analysis for Multi-Horizon Forecasting Paradigms}
\label{appendx:error_bound}

In this section, we provide a mathematical analysis of the upper bounds for three forecasting paradigms: the Direct Paradigm, the Recursive Paradigm \citep{taieb2012review}, and our \method. Let $\mathbf{h}_t \in \mathcal{H}$ denote the underlying hidden state at time $t$. We define the optimal transition dynamics over a length of $\ell$ as $\mathbf{h}_{t+\ell} = f^*(\mathbf{h}_t, \ell)$, and the learned forecasting model as $\hat{\mathbf{h}}_{t+\ell} = f_\theta(\hat{\mathbf{h}}_t, \ell)$. The forecasting error at step $t+\ell$ is measured by the distance $e_{t+\ell} = \|\mathbf{h}_{t+\ell} - \hat{\mathbf{h}}_{t+\ell}\|$. To establish the theoretical bounds, we first introduce two assumptions:

\noindent\textbf{Assumption 1 (Lipschitz Continuous Dynamics).} \textit{We assume the optimal transition function $f^*$ is Lipschitz continuous \citep{asadi2018lipschitz} with respect to the initial state. Specifically, there exists a 1-step Lipschitz constant $\lambda \ge 1$ such that for any states $\mathbf{h}_1, \mathbf{h}_2 \in \mathcal{H}$ and length $\ell \ge 1$, the following holds:}
\begin{equation}
\|f^*(\mathbf{h}_1, \ell) - f^*(\mathbf{h}_2, \ell)\| \le \lambda^\ell \|\mathbf{h}_1 - \mathbf{h}_2\|.
\label{eq:lipschitz}
\end{equation}

\noindent\textbf{Assumption 2 (Strictly Superadditive Approximation Error).} \textit{We assume that for any state $\mathbf{h} \in \mathcal{H}$ within the latent space, the local approximation error of the learned model $f_\theta$ relative to the true dynamics \citep{janner2019trust} $f^*$ is bounded by a strictly superadditive function $\epsilon(\ell)$. Specifically, for any lengths $\ell_1, \ell_2 \ge 1$:}
\begin{equation}
\forall \mathbf{h} \in \mathcal{H}, \quad \|f^*(\mathbf{h}, \ell) - f_\theta(\mathbf{h}, \ell)\| \le \epsilon(\ell), \quad \text{where } \epsilon(\ell_1 + \ell_2) > \epsilon(\ell_1) + \epsilon(\ell_2).
\label{eq:approx_error}
\end{equation}

\textit{Naturally, $\epsilon(\ell)$ is strictly superadditive, as predicting a longer horizon in a single forward pass forces the model to capture a significantly more complex joint distribution, causing the error to grow faster than linearly \citep{bengio2015scheduled}.}

\noindent\textbf{Assumption 3 (Equivalent Residual Bounds).} \textit{We assume the residual dynamics $\mathbf{h}^{\mathrm{res}}$ are bounded by the same Lipschitz constant $\lambda$ and error bound $\epsilon(\cdot)$ as the base dynamics.} 

\subsection{Bound of the Direct Paradigm}
The Direct Paradigm generates all $P$ future steps via a single mapping. The error originates from the approximation difficulty $\epsilon(P)$.

\noindent\textbf{Proposition 1 (Direct Bound).} \textit{The upper bound for the Direct Paradigm at horizon $P$ is given by the approximation error \citep{marcellino2006comparison} of length $P$:}
\begin{equation}
B_{\mathrm{dir}} = \epsilon(P).
\end{equation}
\begin{proof}
In the Direct paradigm, the model predicts the entire horizon $P$ in one forward pass from the ground-truth state $\mathbf{h}_t$. By Assumption 2, $e_P = \|f^*(\mathbf{h}_t, P) - f_\theta(\mathbf{h}_t, P)\| \le \epsilon(P)$.
\end{proof}

\subsection{Bound of the Recursive Paradigm}
The Recursive paradigm predicts the horizon iteratively with a fixed step $\ell=1$. The total horizon $P$ requires $P$ consecutive forward passes \citep{ross2011reduction}, leading to error accumulation.

\noindent\textbf{Proposition 2 (Recursive Bound).} \textit{The upper bound for the Recursive paradigm after $P$ steps is given by:}
\begin{equation}
B_{\mathrm{rec}} = \epsilon(1) \frac{\lambda^P - 1}{\lambda - 1} \quad (\text{assuming } \lambda > 1).
\end{equation}
\begin{proof}
At each step $i$, using the triangle inequality \citep{kreyszig1991introductory} and Eq.~\eqref{eq:lipschitz}:
\begin{align}
e_i &= \|f^*(\mathbf{h}_{t+i-1}, 1) - f_\theta(\hat{\mathbf{h}}_{t+i-1}, 1)\| \\
&\le \|f^*(\mathbf{h}_{t+i-1}, 1) - f^*(\hat{\mathbf{h}}_{t+i-1}, 1)\| + \|f^*(\hat{\mathbf{h}}_{t+i-1}, 1) - f_\theta(\hat{\mathbf{h}}_{t+i-1}, 1)\| \\
&\le \lambda \cdot e_{i-1} + \epsilon(1).
\end{align}
Unrolling this recurrence from $e_0=0$ yields $e_P \le \epsilon(1) \sum_{i=0}^{P-1} \lambda^i = \epsilon(1) \frac{\lambda^P - 1}{\lambda - 1}$.
\end{proof}

\subsection{Bound of the \method}
\method\ fuses a global coarse predictor with an adaptive scheduling branch to generate the final prediction: $\hat{\mathbf{h}}_{t+P} = \hat{\mathbf{h}}_{t+P}^{\mathrm{coarse}} + \alpha \hat{\mathbf{h}}_{t+P}^{\mathrm{sched}}$, where $\alpha \in (0, 1)$ is the learned combination gate. The scheduling branch $\hat{\mathbf{h}}^{\mathrm{sched}}$ predicts the residual dynamics via $K$ adaptive segments of lengths $\{\ell_k\}$. 

\noindent\textbf{Theorem 1 (\method\ Optimal Upper Bound).} \textit{Let $\mathcal{S}_P$ be the set of all valid scheduling partitions such that $\sum_{k=1}^K \ell_k = P$. The optimal upper bound of \method\ is defined as the infimum over all combination gates $\alpha \in (0,1)$ and valid schedules $\boldsymbol{\ell} \in \mathcal{S}_P$:}
\begin{equation}
B_{\mathrm{LeapTS}}^{\star}
=
\inf_{\alpha, \boldsymbol{\ell}} \left[ (1 - \alpha) \epsilon(P) + \alpha \sum_{k=1}^K \lambda^{P - \tau_k} \epsilon(\ell_k) \right],
\end{equation}
\textit{where $\tau_k = \sum_{i=1}^k \ell_i$ denotes the accumulated prediction steps. Consequently, the optimal upper bound of \method\ is no worse than the Direct and Recursive upper bounds:}
\begin{equation}
B_{\mathrm{LeapTS}}^{\star} \le \min(B_{\mathrm{dir}}, B_{\mathrm{rec}}).
\end{equation}

\begin{proof}
By definition, the target residual is $\mathbf{h}^{\mathrm{res}} = \mathbf{h}_{t+P} - \hat{\mathbf{h}}_{t+P}^{\mathrm{coarse}}$. For a specific schedule $\boldsymbol{\ell}$, the total prediction error can be expanded as:
\begin{align}
\|\mathbf{h}_{t+P} - (\hat{\mathbf{h}}_{t+P}^{\mathrm{coarse}} + \alpha \hat{\mathbf{h}}_{t+P}^{\mathrm{sched}})\| 
&= \|\mathbf{h}_{t+P} - \hat{\mathbf{h}}_{t+P}^{\mathrm{coarse}} - \alpha \mathbf{h}^{\mathrm{res}} + \alpha \mathbf{h}^{\mathrm{res}} - \alpha \hat{\mathbf{h}}_{t+P}^{\mathrm{sched}}\| \\
&= \|(1 - \alpha)\mathbf{h}^{\mathrm{res}} + \alpha (\mathbf{h}^{\mathrm{res}} - \hat{\mathbf{h}}_{t+P}^{\mathrm{sched}})\|.
\end{align}
By the triangle inequality:
\begin{equation}
\|\mathbf{h}_{t+P} - (\hat{\mathbf{h}}_{t+P}^{\mathrm{coarse}} + \alpha \hat{\mathbf{h}}_{t+P}^{\mathrm{sched}})\|
\le
(1 - \alpha) \|\mathbf{h}^{\mathrm{res}}\| + \alpha \|\mathbf{h}^{\mathrm{res}} - \hat{\mathbf{h}}_{t+P}^{\mathrm{sched}}\|.
\end{equation}
From Proposition 1, the coarse prediction error is bounded by $\|\mathbf{h}^{\mathrm{res}}\| \le \epsilon(P)$. The second term, $\|\mathbf{h}^{\mathrm{res}} - \hat{\mathbf{h}}_{t+P}^{\mathrm{sched}}\|$, represents the error of the scheduling branch when fitting this residual using $K$ recursive segments. Following the unrolling logic of Proposition 2 with varying segment lengths $\ell_k$, this segment-wise error is bounded by $\sum_{k=1}^K \lambda^{P - \tau_k} \epsilon(\ell_k)$. Taking the limit as $\alpha \to 0$ recovers the Direct upper bound $B_{\mathrm{dir}}$, and taking the limit as $\alpha \to 1$ with $\boldsymbol{\ell}_{\mathrm{rec}} = (1, 1, \dots, 1)$ recovers the Recursive upper bound $B_{\mathrm{rec}}$. Since $B_{\mathrm{LeapTS}}^{\star}$ is defined as the infimum over all valid $(\alpha,\boldsymbol{\ell})$, it follows that $B_{\mathrm{LeapTS}}^{\star} \le \min(B_{\mathrm{dir}}, B_{\mathrm{rec}})$.
\end{proof}

In summary, Theorem 1 demonstrates that \method\ achieves an optimal upper bound no worse than the Direct and Recursive paradigms. It mitigates the Recursive paradigm's exponential drift ($\lambda^P$) by using the global coarse branch as an anchor and reducing jump steps ($K \ll P$). Simultaneously, it overcomes the Direct paradigm's high approximation bias ($\epsilon(P)$) by decomposing the mapping into scheduled segments ($\sum \epsilon(\ell_k) \ll \epsilon(P)$). 

\section{Implementation Details} \label{appendx:imp_detail}

\paragraph{Datasets Description.}
To comprehensively evaluate the proposed framework, experiments are conducted on both real-world and synthetic datasets, covering multivariate and univariate forecasting scenarios. As summarized in Table~\ref{tab:dataset_full}, the real-world datasets span multiple domains, including energy (ETTh1, ETTh2, ETTm1, ETTm2, Electricity), weather (Weather), finance (Exchange Rate, SP500), healthcare (ILI, COVID), web traffic (Wiki), and transportation (PEMS04, PEMS07). For univariate forecasting, the M4 dataset is adopted, which consists of large-scale heterogeneous time series across multiple temporal granularities ranging from hourly to yearly. In addition, we construct synthetic datasets to provide controlled evaluation settings, including two multivariate scenarios and one univariate scenario with consistent configurations.

\textbf{ETT} \citep{zhou2021informer} covers transformer-related multivariate measurements collected from July 2016 to July 2018. The dataset includes variables associated with load conditions and oil temperature. ETTh1 and ETTh2 are organized at the hourly level, while ETTm1 and ETTm2 are recorded every 15 minutes.

\textbf{Weather} \citep{wu2021autoformer} corresponds to meteorological observations collected throughout 2020. It contains 21 environmental variables, including temperature, humidity, and other atmospheric indicators. The data are sampled at 10-minute intervals.

\textbf{Electricity} \citep{wu2021autoformer} spans the period from July 2016 to July 2019 and records electricity consumption at an hourly frequency. The dataset consists of 321 variables, each corresponding to an individual client. Together, these variables form a high-dimensional multivariate time series reflecting client-level consumption patterns.

\textbf{Exchange Rate} \citep{wu2021autoformer} includes daily exchange-rate series over a long historical window from January 1990 to October 2010. Each variable corresponds to one currency-related series. 

\textbf{ILI}\footnote{\url{https://gis.cdc.gov/grasp/fluview/fluportaldashboard.html}} records weekly counts of influenza-like illness patients from 2002 to 2021.

\textbf{SP500} records daily S\&P 500 index data from January 1993 to February 2025. The dataset includes variables such as opening price, closing price, and trading volume. 

\textbf{COVID} \citep{chen2022tamp} consists of daily COVID-19 hospitalization records in California during 2020. 

\textbf{Wiki}\footnote{\url{https://www.kaggle.com/datasets/sandeshbhat/wikipedia-web-traffic-201819}} records daily page-view counts for Wikipedia articles from 2018 to 2019. Each variable corresponds to one Wikipedia page, and the first 99 channels are used in this study.

\textbf{PEMS04} and \textbf{PEMS07} \citep{liu2022scinet} are traffic datasets constructed from road-sensor measurements in California. The observations are aggregated at 5-minute intervals, forming high-frequency traffic flow series. Each variable corresponds to a traffic sensor, describing the temporal evolution of traffic conditions across the network.

\textbf{M4} \citep{makridakis2018m4} is a large-scale univariate time series collection. It contains 100,000 individual series across multiple temporal frequencies, including yearly, quarterly, monthly, weekly, daily, and hourly data. 

\begin{table*}[htbp!]
\caption{Summary of datasets used in this work. We include both real-world and synthetic datasets across multivariate and univariate forecasting tasks.}
\label{tab:dataset_full}
\vskip 0.05in
\centering
\resizebox{\textwidth}{!}{
\begin{threeparttable}
\renewcommand{\arraystretch}{1.2}
\setlength{\tabcolsep}{3pt}

\begin{tabular}{c|c|c|c|c|c|c|c}
\toprule
\textbf{Type} & \textbf{Task} & \textbf{Dataset} & \textbf{Dim} & \textbf{Length} & \textbf{Split (Train/Val/Test, \%)} & \textbf{Freq.} & \textbf{Domain} \\
\toprule

\multirow{19}{*}{Real-world} 
& \multirow{13}{*}{Multivariate} 
& ETTh1 & 7 & 17,420 & 60.0/20.0/20.0 & 1 hour & Temperature \\
& & ETTh2 & 7 & 17,420 & 60.0/20.0/20.0 & 1 hour & Temperature \\
& & ETTm1 & 7 & 69,680 & 60.0/20.0/20.0 & 15 min & Temperature \\
& & ETTm2 & 7 & 69,680 & 60.0/20.0/20.0 & 15 min & Temperature \\
& & Electricity & 321 & 26,304 & 70.0/10.0/20.0 & 1 hour & Electricity \\
& & Exchange & 8 & 7,587 & 70.0/10.0/20.0 & Daily & Finance \\
& & Weather & 21 & 52,696 & 70.0/10.0/20.0 & 10 min & Weather \\
& & ILI & 7 & 966 & 70.0/10.0/20.0 & Weekly & Health \\
& & SP500 & 5 & 8,077 & 70.0/10.0/20.0 & Daily & Finance \\
& & COVID & 55 & 335 & 70.0/10.0/20.0 & Daily & Health \\
& & Wiki & 99 & 730 & 70.0/10.0/20.0 & Daily & Web \\
& & PEMS04 & 307 & 16,992 & 70.0/10.0/20.0 & 5 min & Transportation \\
& & PEMS07 & 883 & 28,224 & 70.0/10.0/20.0 & 5 min & Transportation \\

\cmidrule{2-8}

& \multirow{6}{*}{Univariate} 
& M4-Yearly & 1 & 858,458 & 83.9/0.0/16.1 & Yearly & Demographic \\
& & M4-Quarterly & 1 & 2,406,108 & 92.0/0.0/8.0 & Quarterly & Finance \\
& & M4-Monthly & 1 & 11,246,411 & 92.3/0.0/7.7 & Monthly & Industry \\
& & M4-Weekly & 1 & 371,579 & 98.7/0.0/1.3 & Weekly & Macro \\
& & M4-Daily & 1 & 10,023,836 & 99.4/0.0/0.6 & Daily & Micro \\
& & M4-Hourly & 1 & 373,372 & 94.7/0.0/5.3 & Hourly & Other \\

\midrule
\multirow{3}{*}{Synthetic} 

& \multirow{2}{*}{Multivariate} 
& Scenario 1 & 30 & 20,000 & 60.0/20.0/20.0 & -- & Other \\
& & Scenario 2 & 3 & 20,000 & 60.0/20.0/20.0 & -- & Other \\

\cmidrule{2-8}

& Univariate 
& Scenario 3 & 1 & 20,000 & 60.0/20.0/20.0 & -- & Other \\

\bottomrule
\end{tabular}

\end{threeparttable}
}
\end{table*}

\paragraph{Baseline Details}
To evaluate the effectiveness of \method~in forecasting tasks, we compare it with a diverse set of advanced baselines covering multiple modeling paradigms. Specifically, we include the linear forecasting model DLinear~\citep{zeng2023transformers}; temporal component decomposition methods, including Amplifier~\citep{fei2025amplifier} and TimesNet~\citep{wu2023timesnet}; multi-scale modeling methods, including AMD~\citep{hu2025adaptive} and TimeMixer~\citep{wang2025timemixer++}; MLP-based methods, including TSMixer~\citep{chen2023tsmixer}; the patch-based model PatchTST~\citep{Yuqietal-2023-PatchTST}; and the Transformer-based model iTransformer~\citep{liu2024itransformer}. These baselines cover a wide range of model design paradigms and provide a comprehensive basis for evaluation.

\paragraph{Metric Details.}
We evaluate the forecasting performance using Mean Squared Error (MSE) and Mean Absolute Error (MAE) for real-world datasets. For the M4 benchmark \citep{makridakis2018m4}, we follow the official evaluation protocol of the M4 competition and adopt Symmetric Mean Absolute Percentage Error (SMAPE), Mean Absolute Scaled Error (MASE), and Overall Weighted Average (OWA), where OWA is the primary metric used in the M4 competition. The metrics are defined as:
\begin{alignat}{2}
\text{MSE} &= \frac{1}{P} \sum_{i=1}^{P} (\mathbf{Y}_{i} - \widehat{\mathbf{Y}}_{i})^2, & \hspace{1.5em} \text{MAE} &= \frac{1}{P} \sum_{i=1}^{P}|\mathbf{Y}_{i} - \widehat{\mathbf{Y}}_{i}|, \\
\text{SMAPE} &= \frac{200}{P} \sum_{i=1}^{P} \frac{|\mathbf{Y}_{i} - \widehat{\mathbf{Y}}_{i}|}{|\mathbf{Y}_{i}| + |\widehat{\mathbf{Y}}_{i}|}, & \hspace{1.5em} \text{OWA} &= \frac{1}{2} \left[ \frac{\text{SMAPE}}{\text{SMAPE}_{\textrm{Na\"ive2}}} + \frac{\text{MASE}}{\text{MASE}_{\textrm{Na\"ive2}}} \right], \\
\text{MAPE} &= \frac{100}{P} \sum_{i=1}^{P} \frac{|\mathbf{Y}_{i} - \widehat{\mathbf{Y}}_{i}|}{|\mathbf{Y}_{i}|}, & \hspace{1.5em} \text{MASE} &= \frac{\frac{1}{P} \sum_{i=1}^{P} |\mathbf{Y}_{i} - \widehat{\mathbf{Y}}_{i}|}{\frac{1}{L-s}\sum_{j=s+1}^{L}|\mathbf{X}_j - \mathbf{X}_{j-s}|}.
\end{alignat}

where $s$ is the periodicity of the data. $\mathbf{X}\in\mathbb{R}^{L\times N}$ denotes the historical observations with look-back length $L$. $\mathbf{Y},\widehat{\mathbf{Y}}\in\mathbb{R}^{P\times N}$ denote the ground truth and prediction with $P$ future time points and $N$ variables. $\mathbf{Y}_{i}$ denotes the $i$-th future time step, and $\mathbf{X}_{j}$ denotes the $j$-th historical time step.

\paragraph{Experiment Details.}
All experiments were implemented in PyTorch~\citep{paszke2019pytorch} and conducted on multiple NVIDIA A100 80GB GPUs, with each experiment repeated five times. The learning rate was set in the range from $2.5\times10^{-4}$ to $10^{-2}$, and the ADAM optimizer \citep{kingma2014adam} was used for optimization. The Huber loss was adopted during training \citep{huber1992robust}. For model configuration, the number of hidden layers for both the \textsc{H-Controller} and the prediction head was set from 1 to 3, and the hidden dimension was selected from 1 to 256. Specifically, for low-dimensional datasets such as ETTh1, smaller hidden dimensions were used. In contrast, for high-dimensional datasets such as Electricity, larger dimensions were adopted to capture more complex dependencies. For baseline comparisons, results were directly reported from the original papers when the experimental settings were consistent \citep{wang2024timemixer}. Otherwise, results were reproduced based on the benchmark framework of the Time-Series Library\footnote{\url{https://github.com/thuml/Time-Series-Library}}.

\paragraph{Hyperparameter Tuning on Synthetic Datasets.} 
To ensure a fair evaluation, we conduct a grid search to find the optimal configurations. For \method, we search the hidden dimension in $\{10, 20, 30, 40\}$ and the number of layers in $\{2, 3, 4\}$. For TimeMixer and PatchTST, we expand the search space of hidden dimensions to $\{10, 20, 30, 40, 128, 256\}$ and search the number of encoder/decoder layers in $\{2, 3, 4\}$. We report the best performance for all models.

\section{Length Ranges for Different Scales}
\label{appendx:length_ranges}

As introduced in Section \ref{step_section}, the \textsc{H-Controller} decomposes the segment length selection into a two-level hierarchical decision. Given the controller state $\mathbf{h}_k$ at step $k$, the model first selects a temporal scale category $c \in \{\text{short}, \text{mid}, \text{long}\}$ via a categorical routing module. Conditioned on the selected scale, a category-specific regression head then predicts a segment length within a predefined interval. This section details the construction of these scale-specific boundaries and their valid ranges.

\subsection{Scale Anchors}

The boundaries of the length intervals dynamically adapt to the look-back length $L$ and the forecasting horizon $P$. Intuitively, a longer historical context provides richer information, enabling the model to confidently take larger advancement steps. Formally, the three upper-bound anchors are defined as:
\begin{equation}
    L_{\text{short}} = \max(1, \min(\lfloor L/4 \rfloor, P-1))
\end{equation}
\begin{equation}
    L_{\text{mid}} = \max(L_{\text{short}} + 1, \min(\lfloor L/2 \rfloor, P-1))
\end{equation}
\begin{equation}
    L_{\text{long}} = P
\end{equation}

By anchoring the short and medium scales to fractions of $L$, the step size naturally scales with the available historical evidence. Furthermore, setting $L_{\text{long}} = P$ allows the model to predict simple patterns in a single forward pass, thereby avoiding the inference overhead of multi-round scheduling. It is worth noting that if $P \le \lfloor L/4 \rfloor + 1$, the model skips these anchors and directly uses a single-level decision.

\subsection{Category-Specific Intervals and Length Mapping}

Based on the derived anchors, we construct the valid length interval $[L_c^{\min}, L_c^{\max}]$ for each category $c$. The maximum boundaries are directly set to the anchor values ($L_c^{\max} = L_c$), and the minimum boundary for the short scale is fixed as $L_{\text{short}}^{\min} = 1$. To prevent excessive overlap and ensure progressive scaling, the remaining minimum boundaries are defined as:
\begin{align}
     L_{\text{mid}}^{\min} &= \max\big(2, \min(L_{\text{mid}}^{\max}, L_{\text{short}}^{\max} + 1)\big), \\
L_{\text{long}}^{\min} &= \max\big(3, \min(L_{\text{long}}^{\max}, \max(L_{\text{mid}}^{\max} + 1, \lfloor L_{\text{long}}^{\max}/2 \rfloor))\big).  
\end{align}
 
During the low-level scheduling phase, the raw output from the category-specific length head $f^{\text{len}}_c(\mathbf{h}_k)$ is mapped into this defined interval via a sigmoid activation: $\ell_{k,c} = L_c^{\min} + (L_c^{\max} - L_c^{\min}) \cdot \sigma\big(f^{\text{len}}_c(\mathbf{h}_k)\big)$. The continuous length of the selected category is subsequently clipped by the remaining forecasting horizon and rounded to the nearest integer to yield the final execution length $\ell_k$.

\section{Synthetic Dataset Details}
\label{appendx:Syntheti}

Although real-world benchmarks evaluate overall forecasting accuracy, they often mix various complex factors, making it difficult to clearly assess a model's specific architectural strengths. To explicitly isolate and test distinct behaviors under specific temporal complexities, we generate three synthetic datasets, each consisting of 20,000 time steps.

\textbf{Scenario 1: High-Dimensional Heterogeneous Series.} This 30-dimensional dataset ($\Delta t = 0.05$) contains three latent clusters (10 variables each) to evaluate the decoupling of heterogeneous temporal patterns:
\begin{itemize}
\vspace{-5pt}
    \setlength{\itemsep}{2pt} 
    \item \textbf{Cluster A (Continuous Periodic Patterns):} Generated via the Van der Pol system \citep{lasalle1949relaxation} to simulate stable, non-linear seasonal fluctuations. $dy_1/dt = y_2$ and $dy_2/dt = \mu (1 - y_1^2) y_2 - y_1$ ($\mu = 2.0$), with observation noise $\mathcal{N}(0, 0.05^2)$.
    \item \textbf{Cluster B (Sudden Sharp Spikes):} Generated via the FitzHugh-Nagumo model \citep{fitzhugh1961impulses} to simulate event-driven anomalies with sharp, rapid state changes. $dv/dt = v - v^3/3 - w + I(t)$ and $dw/dt = 0.08(v + 0.7 - 0.8w)$, where the intermittent pulse $I(t) = 0.5$ if $(t \bmod 50) < 2.0$, and $0$ otherwise.
    \item \textbf{Cluster C (Sharp Jumps with Slow Recovery):} Generated via a damped harmonic oscillator \citep{strogatz2024nonlinear} to simulate time series with decaying trends. $dy_1/dt = y_2$ and $dy_2/dt = -0.15 y_2 - 1.0 y_1$. The underlying states are periodically reset every 2,000 steps from $\mathcal{U}(-2, 2)$ to simulate repeated external shocks.
\end{itemize}

\textbf{Scenario 2: Causally Driven Multivariate Series.} This 3-dimensional system ($X, Y, U$, $\Delta t = 0.05$) simulates a multi-variable time series exhibiting severe non-stationary regime shifts. The core dynamics follow the classic Brusselator equations \citep{prigogine1968symmetry}, which are causally modulated by a slow-moving external driver $U$ acting as an unobserved macro-trend. This exogenous driver smoothly tracks a periodic target: $dU/dt = 0.2 (U_{\text{target}}(t) - U)$, where $U_{\text{target}}(t) = 1.0 + 0.5 \sin(0.1t)$. This hidden factor continuously drives the interacting variables ($X, Y$):
\begin{equation}
    \frac{dX}{dt} = 1.0 + X^2Y - (B(t) + 1)X, \quad \frac{dY}{dt} = B(t)X - X^2Y
\end{equation}

where the time-varying coefficient is dynamically governed by $B(t) = 2.5 + 2.0 U(t)$. By continuously modulating $B(t)$, the unobserved driver forces the sequence to alternate unpredictably between flat, low-volatility baseline periods and explosive, high-frequency fluctuations.

\textbf{Scenario 3: Univariate Series Driven by Hidden Variables.} This dataset tests whether the model can forecast a 1D sequence when its underlying driving factors are unobserved ($\Delta t = 0.02$). We generate the data using the classic Lorenz system \citep{lorenz2017deterministic}, which produces complex, non-repeating patterns:
\begin{equation}
    \frac{dx}{dt} = \sigma(y - x), \quad \frac{dy}{dt} = x(\rho - z) - y, \quad \frac{dz}{dt} = xy - \beta z
\end{equation}

using standard parameters ($\sigma = 10.0, \rho = 28.0, \beta = 8/3$). To create a challenging forecasting task, we intentionally hide the interacting variables $x$ and $y$. By keeping only the $z$ sequence, we force the model to infer the missing hidden states and predict future values purely based on the historical movements of $z$.

\section{Signal Decomposition Mechanism}
\label{app:decomposition}

This section details the explicit decomposition of the \textsc{H-Controller}'s internal state update. At each step $k$, the model builds a control signal $\mathbf{u}_k$ based on the current context: $\mathbf{u}_k = \tanh\!\big(\mathbf{W}_{u} [\rho_k \;\Vert\; \bar{\ell}_{k-1} \;\Vert\; \boldsymbol{\pi}_{k-1} \;\Vert\; \mathbf{c}_{k-1}]\big)$. This context vector combines the remaining horizon, previous step length, previous category, and segment summary. We split the controller's state update into two parts: one driven by control signal changes ($\Delta\mathbf{u}_k$) and one driven by temporal progress ($\Delta\tau_k$):
\begin{equation}
    \mathbf{h}_{k+1} = \mathbf{h}_k + \underbrace{\mathbf{F}(\mathbf{h}_k, \mathbf{u}_k) \Delta\mathbf{u}_k}_{\Delta\mathbf{h}_k^{\mathrm{ctrl}}} + \underbrace{\mathbf{G}(\mathbf{h}_k, \mathbf{u}_k) \Delta\tau_k}_{\Delta\mathbf{h}_k^{\mathrm{time}}}
\end{equation}

where $\mathbf{F}$ and $\mathbf{G}$ are the learned networks for the control and temporal components. To prevent positive and negative values from canceling each other out, we sum their absolute values: $C_k = \sum_j |\Delta h_{k,j}^{\mathrm{ctrl}}|$ and $T_k = \sum_j |\Delta h_{k,j}^{\mathrm{time}}|$. We then calculate their relative ratios: $\eta_k^{\mathrm{ctrl}} = C_k / (C_k + T_k + \varepsilon)$ and $\eta_k^{\mathrm{time}} = T_k / (C_k + T_k + \varepsilon)$. These ratios clearly show whether a step is driven more by the control signal ($\eta_k^{\mathrm{ctrl}} \to 1$) or by the temporal signal ($\eta_k^{\mathrm{time}} \to 1$).

\section{Ablation Analysis}
\label{appendx:ablation}

\textbf{Ablation on the Scheduling Trace.} To verify the effectiveness of the scheduling trace, we compare it against two representative baselines: 

1) \textbf{Monte Carlo Simulation}: This approach generates random scheduling traces by arbitrarily partitioning the prediction horizon. The final performance is then calculated by averaging the forecasting errors across all sampled traces. To manage computational costs, we sample $N=10$ random traces per input window for the high-dimensional Electricity and PEMS04 datasets, and $N=100$ traces for all other datasets. 

2) \textbf{Fixed Period Step}: The forecasting process is executed using a fixed step size determined by the dataset's inherent periodicity (e.g., a step of 24 for hourly data).
\begin{table}[htbp!] 
\caption{Ablation study on the scheduling trace. We report the MAE metric. 
The \best{best} and \second{second-best} performances are highlighted. 
\textbf{\textit{Average}} denotes the mean performance across all datasets, and \textbf{Improvement} indicates the relative MAE reduction of \method\ 
with respect to the \textbf{strongest baseline}.}
\label{tab:ablation_trace}
\vskip 0.1in
\centering
\begin{threeparttable}
\footnotesize 
\renewcommand{\arraystretch}{1.0} 
\setlength{\tabcolsep}{4pt} 

\newcommand{\num}[1]{{#1}} 
\newcommand{\met}[1]{{\textbf{#1}}}
\newcommand{\modelhead}[1]{#1}
\newcommand{\yearhead}[1]{\scalebox{0.9}{#1}}

\begin{tabular}{@{}c|ccc|c@{}}
\toprule
\multirow{2}{*}{\textbf{Dataset}} & 
\multicolumn{1}{c}{\modelhead{\textbf{\method}}} & 
\multicolumn{1}{c}{\modelhead{\textbf{Monte Carlo}}} & 
\multicolumn{1}{c|}{\modelhead{\textbf{Fixed Step}}} & 
\multirow{2}{*}{\textbf{\textit{Improvement}}} \\

& 
\multicolumn{1}{c}{\yearhead{(Ours)}} & 
\multicolumn{1}{c}{\yearhead{(Baseline 1)}} & 
\multicolumn{1}{c|}{\yearhead{(Baseline 2)}} & \\
\midrule

\textit{\textbf{Electricity}} & \best{\num{0.2260}} & \num{0.2773} & \second{\num{0.2437}} & \textbf{\textit{\num{7.26\%}}} \\
\textit{\textbf{ETTm1}}       & \best{\num{0.3392}} & \num{0.3696} & \second{\num{0.3652}} & \textbf{\textit{\num{7.12\%}}} \\
\textit{\textbf{ETTm2}}       & \best{\num{0.2354}} & \num{0.2405} & \second{\num{0.2399}} & \textbf{\textit{\num{1.88\%}}} \\
\textit{\textbf{Exchange}}    & \best{\num{0.1591}} & \second{\num{0.1752}} & \num{0.1793} & \textbf{\textit{\num{9.19\%}}} \\
\textit{\textbf{ILI}}         & \best{\num{0.8371}} & \num{0.9748} & \second{\num{0.9527}} & \textbf{\textit{\num{12.13\%}}} \\
\textit{\textbf{PEMS04}}      & \best{\num{0.2872}} & \second{\num{0.4126}} & \num{0.4135} & \textbf{\textit{\num{30.39\%}}} \\
\textit{\textbf{Weather}}     & \best{\num{0.1778}} & \num{0.1979} & \second{\num{0.1959}} & \textbf{\textit{\num{9.24\%}}} \\

\midrule
\rowcolor{rowblue}
\textit{\textbf{Average}} & \best{\num{0.3231}} & \num{0.3783} & \second{\num{0.3700}} & \textbf{\textit{\num{11.03\%}}} \\
\bottomrule
\end{tabular}
\end{threeparttable}
\end{table}

As summarized in Table~\ref{tab:ablation_trace}, \method\ consistently outperforms both the stochastic and fixed-step baselines across all evaluated datasets. Notably, \method\ achieves substantial performance gains, with relative improvements reaching up to 30.39\% on the PEMS04 dataset and 12.13\% on the ILI dataset. These results demonstrate the effectiveness of our proposed scheduling method.

\textbf{Ablation on Hierarchical Scheduling.} To evaluate the effectiveness of hierarchical scheduling, we compare \method\ with a variant (\textit{w/o High-Level}) that directly regresses the advancement length without high-level scale selection. As shown in Table~\ref{tab:ablation_hierarchical}, removing the high-level scale selection leads to a consistent performance drop across all evaluated datasets.
\begin{table}[htbp!] 
  \caption{Ablation study on the hierarchical scheduling mechanism. We compare the full \method\ with a variant (\textit{w/o High-Level}) that directly predicts the step length without the high-level scale selection. The forecasting horizon is set to $P=36$ for ILI and $P=60$ for others. \textbf{\textit{Average}} denotes the mean performance. The \best{best} performances are highlighted.}
  \label{tab:ablation_hierarchical}
  \vskip 0.1in
  \centering
  \begin{threeparttable}
  \small 
  \renewcommand{\arraystretch}{1.1} 
  \setlength{\tabcolsep}{10pt} 

  \begin{tabular}{l|cc|cc}
    \toprule
    \multirow{2}{*}{\textbf{Dataset}} & \multicolumn{2}{c|}{\textbf{\method} \textit{(Ours)}} & \multicolumn{2}{c}{\textbf{\textit{w/o High-Level}}} \\
    \cmidrule(lr){2-3} \cmidrule(lr){4-5}
    & \textbf{MSE} & \textbf{MAE} & \textbf{MSE} & \textbf{MAE} \\
    \midrule

    \textit{\textbf{Weather}}     & \best{0.1396} & \best{0.1778} & 0.1404 & 0.1796 \\
    \textit{\textbf{Electricity}} & \best{0.1314} & \best{0.2260} & 0.1327 & 0.2279 \\
    \textit{\textbf{ETTh1}}       & \best{0.3511} & \best{0.3736} & 0.3525 & 0.3766 \\
    \textit{\textbf{ETTh2}}       & \best{0.2458} & \best{0.3091} & 0.2462 & 0.3106 \\
    \textit{\textbf{ETTm1}}       & \best{0.2978} & \best{0.3392} & 0.2991 & 0.3412 \\
    \textit{\textbf{ETTm2}}       & \best{0.1469} & \best{0.2354} & 0.1479 & 0.2368 \\
    \textit{\textbf{Exchange}}    & \best{0.0527} & \best{0.1591} & 0.0533 & 0.1605 \\
    \textit{\textbf{ILI}}         & \best{1.9480} & \best{0.8371} & 2.1337 & 0.8833 \\

    \midrule
    \rowcolor{rowblue}
    \textit{\textbf{Average}}     & \best{0.4142} & \best{0.3322} & 0.4382 & 0.3395 \\
    \bottomrule
  \end{tabular}
  \end{threeparttable}
\end{table}

\section{Hyperparameter Sensitivity}
\label{appendx:sensitive}

This section investigates the impact of key hyperparameters on the forecasting performance of \method, specifically focusing on the number of hidden layers and the input look-back length. All experiments are conducted on the ETTm1 dataset.

\textbf{Impact of Hidden Layers.} We evaluate the influence of the \textsc{H-Controller}'s depth by varying the number of hidden layers from 1 to 4 while fixing the input length at $L=96$. As summarized in Table~\ref{tab:sensitive_layers}, the forecasting performance of \method\ remains robust across different layer configurations, with only marginal fluctuations in error metrics. 
\begin{table}[htbp!]
  \caption{Sensitivity analysis of the number of hidden layers on ETTm1 dataset ($L=96$). \textbf{Average} denotes the mean performance across all prediction lengths $P$.}
  \label{tab:sensitive_layers}
  \vskip 0.1in
  \centering
  \begin{small}
  \begin{tabular}{c|cc|cc|cc|cc} 
    \toprule
    \textbf{Layers} & \multicolumn{2}{c|}{\textit{\textbf{Layer = 1}}} & \multicolumn{2}{c|}{\textit{\textbf{Layer = 2}}} & \multicolumn{2}{c|}{\textit{\textbf{Layer = 3}}} & \multicolumn{2}{c}{\textit{\textbf{Layer = 4}}} \\
    \cmidrule(lr){1-1} \cmidrule(lr){2-3} \cmidrule(lr){4-5} \cmidrule(lr){6-7} \cmidrule(lr){8-9} 
    \textbf{Horizon ($P$)} & \textbf{MSE} & \textbf{MAE} & \textbf{MSE} & \textbf{MAE} & \textbf{MSE} & \textbf{MAE} & \textbf{MSE} & \textbf{MAE} \\
    \midrule
    18 & 0.1811 & 0.2608 & 0.1769 & 0.2571 & 0.1763 & 0.2575 & 0.1765 & 0.2596 \\
    24 & 0.2152 & 0.2846 & 0.2100 & 0.2812 & 0.2113 & 0.2787 & 0.2102 & 0.2785 \\
    36 & 0.2645 & 0.3170 & 0.2546 & 0.3112 & 0.2519 & 0.3073 & 0.2535 & 0.3081 \\
    48 & 0.2900 & 0.3346 & 0.2821 & 0.3303 & 0.2856 & 0.3359 & 0.2818 & 0.3296 \\
    60 & 0.2978 & 0.3392 & 0.2937 & 0.3336 & 0.2971 & 0.3383 & 0.3062 & 0.3431 \\
    \midrule
    \rowcolor{rowblue} \textit{\textbf{Average}} & \textbf{0.2497} & \textbf{0.3072} & \textbf{0.2435} & \textbf{0.3027} & \textbf{0.2444} & \textbf{0.3036} & \textbf{0.2456} & \textbf{0.3038} \\
    \bottomrule
  \end{tabular}
  \end{small}
\end{table}

\textbf{Impact of Input Length.} We further evaluate the sensitivity of \method\ to various look-back windows $L \in \{96, 192, 336, 512\}$. As summarized in Table~\ref{tab:sensitive_input}, increasing the input length from 96 to 336 generally reduces the forecasting error since the model benefits from a richer historical context. The performance reaches its peak at $L=336$ with an average MSE of 0.2214.

\begin{table}[htbp]
  \caption{Sensitivity analysis of input look-back lengths ($L$) on ETTm1 dataset. \textbf{Average} denotes the mean performance across all prediction lengths $P$.}
  \label{tab:sensitive_input}
  \vskip 0.1in
  \centering
  \begin{small}
  \begin{tabular}{c|cc|cc|cc|cc} 
    \toprule
    \textbf{Input $L$} & \multicolumn{2}{c|}{\textit{\textbf{$L$ = 96}}} & \multicolumn{2}{c|}{\textit{\textbf{$L$ = 192}}} & \multicolumn{2}{c|}{\textit{\textbf{$L$ = 336}}} & \multicolumn{2}{c}{\textit{\textbf{$L$ = 512}}} \\
    \cmidrule(lr){1-1} \cmidrule(lr){2-3} \cmidrule(lr){4-5} \cmidrule(lr){6-7} \cmidrule(lr){8-9} 
    \textbf{Horizon ($P$)} & \textbf{MSE} & \textbf{MAE} & \textbf{MSE} & \textbf{MAE} & \textbf{MSE} & \textbf{MAE} & \textbf{MSE} & \textbf{MAE} \\
    \midrule
    18 & 0.1769 & 0.2571 & 0.1631 & 0.2477 & 0.1605 & 0.2487 & 0.1599 & 0.2476 \\
    24 & 0.2100 & 0.2812 & 0.1933 & 0.2693 & 0.1919 & 0.2708 & 0.2002 & 0.2743 \\
    36 & 0.2546 & 0.3112 & 0.2319 & 0.2959 & 0.2300 & 0.2976 & 0.2354 & 0.2984 \\
    48 & 0.2821 & 0.3303 & 0.2567 & 0.3126 & 0.2547 & 0.3144 & 0.2599 & 0.3155 \\
    60 & 0.2978 & 0.3392 & 0.2688 & 0.3212 & 0.2701 & 0.3238 & 0.2781 & 0.3293 \\
    \midrule
    \rowcolor{rowblue} \textit{\textbf{Average}} & \textbf{0.2443} & \textbf{0.3038} & \textbf{0.2228} & \textbf{0.2893} & \textbf{0.2214} & \textbf{0.2910} & \textbf{0.2267} & \textbf{0.2930} \\
    \bottomrule
  \end{tabular}
  \end{small}
\end{table}

\textbf{Impact of Variable Clusters.} We evaluate the cluster count $G \in \{1, 3, 5, 7\}$ on the ETTm1 dataset. As Table~\ref{tab:sensitive_clusters} shows, performance remains highly stable across all configurations. Notably, a single globally shared controller ($G=1$) achieves the lowest average error. This demonstrates that for low-dimensional time series, sharing a single NCDE controller is entirely sufficient to yield optimal forecasting results while maximizing parameter efficiency.

\begin{table}[htbp]
  \caption{Sensitivity analysis of the number of variable clusters ($G$) on the ETTm1 dataset. \textbf{Average} denotes the mean performance across all prediction lengths $P$.}
  \label{tab:sensitive_clusters}
  \vskip 0.1in
  \centering
  \begin{small}
  \begin{tabular}{c|cc|cc|cc|cc} 
    \toprule
    \textbf{Clusters $G$} & \multicolumn{2}{c|}{\textit{\textbf{$G$ = 1}}} & \multicolumn{2}{c|}{\textit{\textbf{$G$ = 3}}} & \multicolumn{2}{c|}{\textit{\textbf{$G$ = 5}}} & \multicolumn{2}{c}{\textit{\textbf{$G$ = 7}}} \\
    \cmidrule(lr){1-1} \cmidrule(lr){2-3} \cmidrule(lr){4-5} \cmidrule(lr){6-7} \cmidrule(lr){8-9} 
    \textbf{Horizon ($P$)} & \textbf{MSE} & \textbf{MAE} & \textbf{MSE} & \textbf{MAE} & \textbf{MSE} & \textbf{MAE} & \textbf{MSE} & \textbf{MAE} \\
    \midrule
    18 & 0.1776 & 0.2560 & 0.1768 & 0.2593 & 0.1787 & 0.2589 & 0.1767 & 0.2580 \\
    24 & 0.2089 & 0.2790 & 0.2106 & 0.2824 & 0.2116 & 0.2814 & 0.2139 & 0.2845 \\
    36 & 0.2530 & 0.3073 & 0.2535 & 0.3072 & 0.2556 & 0.3112 & 0.2585 & 0.3112 \\
    48 & 0.2788 & 0.3247 & 0.2801 & 0.3284 & 0.2828 & 0.3292 & 0.2840 & 0.3280 \\
    60 & 0.2960 & 0.3368 & 0.2989 & 0.3378 & 0.2987 & 0.3396 & 0.2961 & 0.3382 \\
    \midrule
    \rowcolor{rowblue} \textit{\textbf{Average}} & \textbf{0.2429} & \textbf{0.3008} & \textbf{0.2440} & \textbf{0.3030} & \textbf{0.2455} & \textbf{0.3041} & \textbf{0.2458} & \textbf{0.3040} \\
    \bottomrule
  \end{tabular}
  \end{small}
\end{table}

\section{Forecasting Stability Across Random Seeds}
\label{appendx:error_bars}

To validate the robustness of our framework, we repeated all training and evaluation procedures across 5 independent random seeds, which were sampled from the range $[1, 10000]$. We report the average performance and the corresponding error bars at the 90\%, 95\%, and 99\% confidence levels on the Weather, Electricity, and ETT datasets, as summarized in Table~\ref{tab:errorbar_main}. The results confirm that \method~remains highly stable across different random initializations.

\begin{table}[htbp]
  \caption{Error bars for \method~on six real-world datasets across 90\%, 95\%, and 99\% confidence levels.}\label{tab:errorbar_main}
  \vskip 0.05in
  \centering
  \resizebox{\textwidth}{!}{ 
  \begin{threeparttable}
  \begin{small}
  \renewcommand{\multirowsetup}{\centering}
  \setlength{\tabcolsep}{8pt} 
  \begin{tabular}{l|cc|cc|cc}
    \toprule
    \multirow{2}{*}{\textbf{Dataset}} & \multicolumn{2}{c|}{\textbf{90\% Confidence Level}} & \multicolumn{2}{c|}{\textbf{95\% Confidence Level}} & \multicolumn{2}{c}{\textbf{99\% Confidence Level}} \\
    \cmidrule(lr){2-3}\cmidrule(lr){4-5}\cmidrule(lr){6-7}
    & \textbf{MSE} & \textbf{MAE} & \textbf{MSE} & \textbf{MAE} & \textbf{MSE} & \textbf{MAE} \\
    \midrule
    \textit{Weather}     & $0.113 \pm 0.001$ & $0.144 \pm 0.001$ & $0.113 \pm 0.001$ & $0.144 \pm 0.002$ & $0.113 \pm 0.002$ & $0.144 \pm 0.004$ \\
    \textit{Electricity} & $0.118 \pm 0.000$ & $0.213 \pm 0.000$ & $0.118 \pm 0.000$ & $0.213 \pm 0.000$ & $0.118 \pm 0.001$ & $0.213 \pm 0.001$ \\
    \textit{ETTh1}       & $0.316 \pm 0.000$ & $0.354 \pm 0.000$ & $0.316 \pm 0.000$ & $0.354 \pm 0.000$ & $0.316 \pm 0.001$ & $0.354 \pm 0.000$ \\
    \textit{ETTh2}       & $0.198 \pm 0.002$ & $0.277 \pm 0.002$ & $0.198 \pm 0.003$ & $0.277 \pm 0.003$ & $0.198 \pm 0.004$ & $0.277 \pm 0.005$ \\
    \textit{ETTm1}       & $0.243 \pm 0.002$ & $0.302 \pm 0.002$ & $0.243 \pm 0.003$ & $0.302 \pm 0.002$ & $0.243 \pm 0.005$ & $0.302 \pm 0.004$ \\
    \textit{ETTm2}       & $0.118 \pm 0.000$ & $0.209 \pm 0.000$ & $0.118 \pm 0.000$ & $0.209 \pm 0.001$ & $0.118 \pm 0.001$ & $0.209 \pm 0.001$ \\
    \bottomrule
  \end{tabular}
  \end{small}
  \end{threeparttable}
  }
\end{table}

\section{Showcases}
\label{appendx:showcases}

In this section, we provide additional visualizations to further illustrate the effectiveness of \method. We begin by comparing our paradigm with the direct mapping baseline. Next, we present the forecasting behavior and corresponding latent state evolution under controlled synthetic scenarios. Subsequently, we showcase extensive real-world prediction cases across diverse domains, including Electricity, ETTh1, PEMS04, and Weather. These visual comparisons explicitly demonstrate how \method~dynamically adapts its scheduling segments to capture complex temporal patterns.
\begin{figure*}[htbp]
    \centering
    \includegraphics[width=0.8\linewidth]{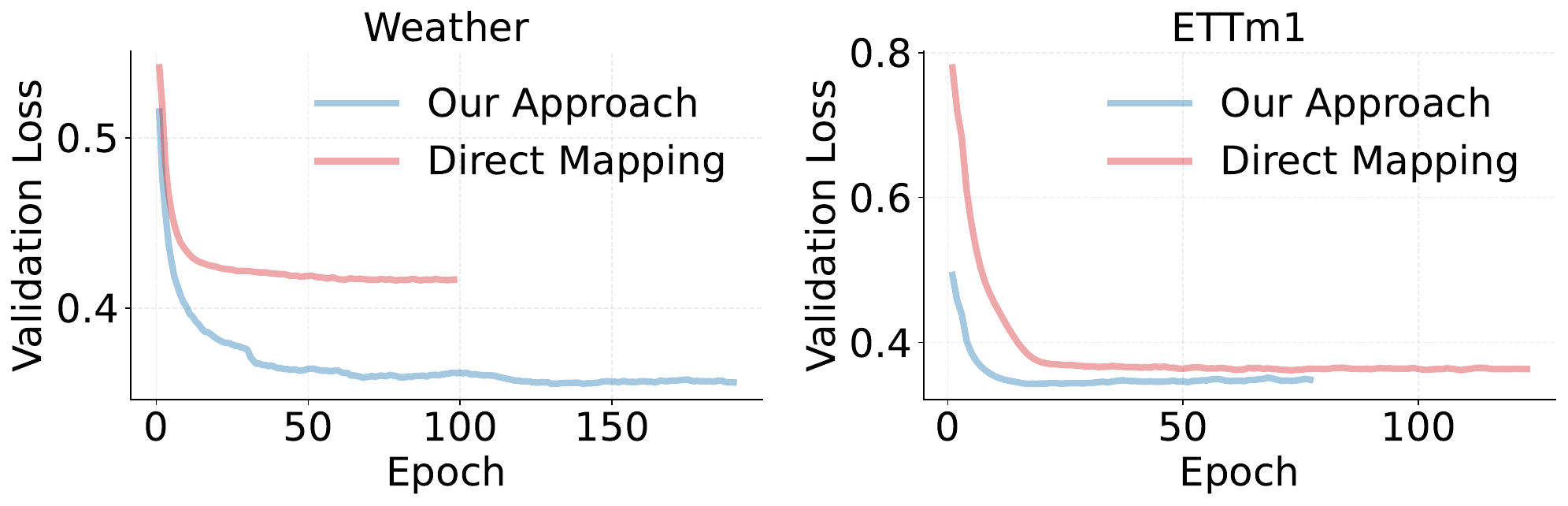}
    \caption{Validation loss on the ETTm1 and Weather datasets. Under the same hyperparameter settings, \method\ converges faster and reaches a lower final loss.}
    \label{loss}
\end{figure*}

\begin{figure*}[htbp]
    \centering
    \includegraphics[width=1\linewidth]{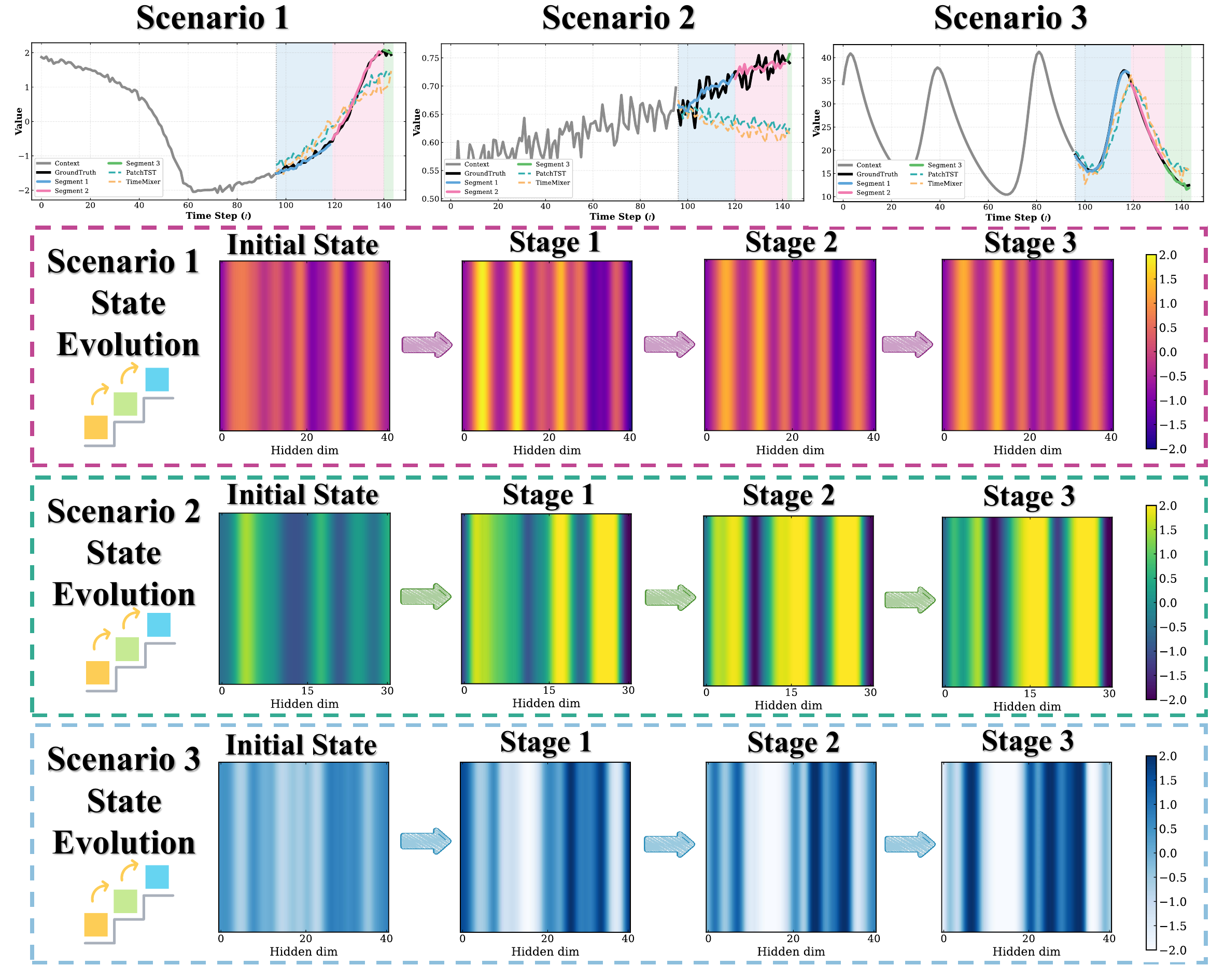}
    \caption{Visualization of forecasting behavior and latent state evolution. The top row shows the predictions across three synthetic scenarios, where different background colors indicate distinct scheduling segments. The bottom rows display the corresponding evolution of the hidden states across successive scheduling stages.}
    \label{decision_fig3}
\end{figure*}

\begin{figure*}[htbp]
    \centering
    \includegraphics[width=1\linewidth]{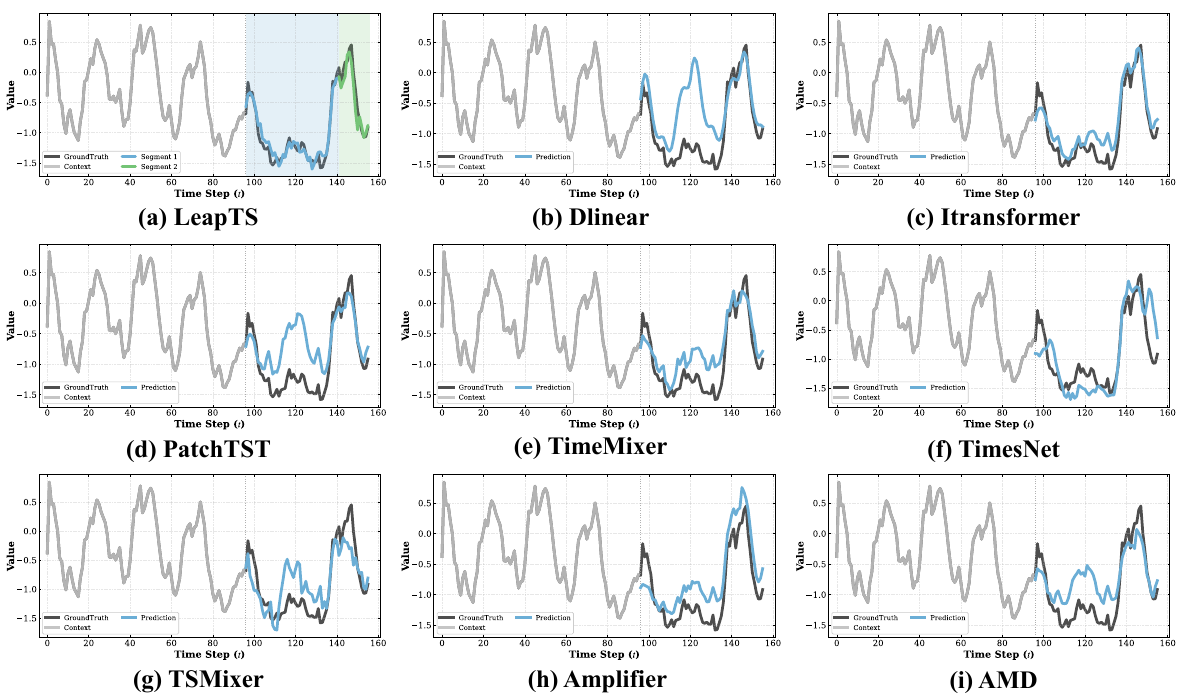}
    \caption{Visualization of forecasting cases from Electricity by different models under the input-96-predict-60 setting. Notably, for \method, both the background regions and the predicted line segments are highlighted with various colors to explicitly illustrate the distinct dynamic scheduling segments.}
    \label{ecl}
\end{figure*}

\begin{figure*}[htbp]
    \centering
    \includegraphics[width=1\linewidth]{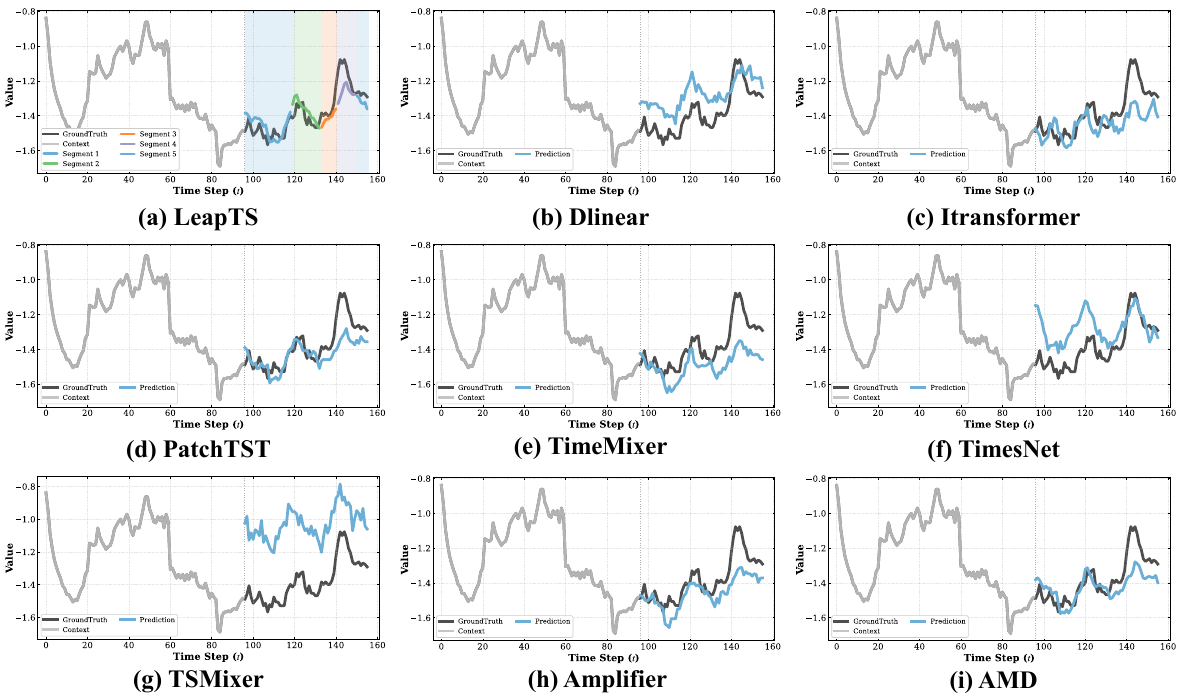}
    \caption{Visualization of forecasting cases from ETTh1 by different models under the input-96-predict-60 setting. Notably, for \method, both the background regions and the predicted line segments are highlighted with various colors to explicitly illustrate the distinct dynamic scheduling segments.}
    \label{etth1}
\end{figure*}

\begin{figure*}[htbp]
    \centering
    \includegraphics[width=1\linewidth]{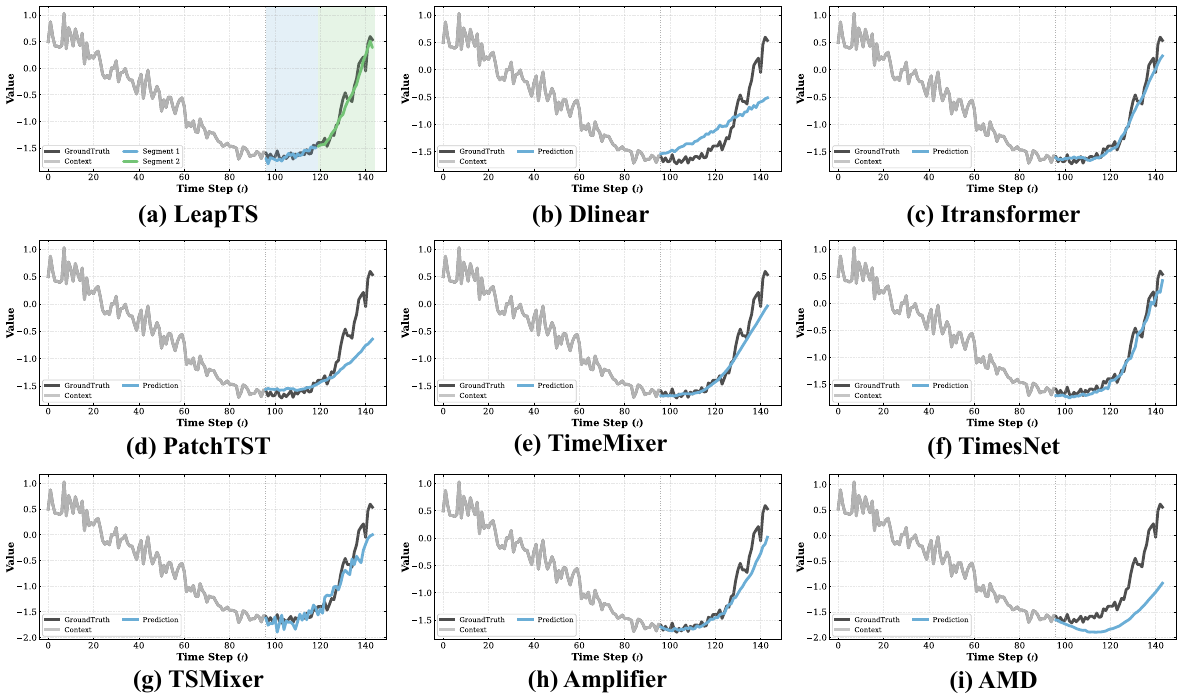}
    \caption{Visualization of forecasting cases from PEMS04 by different models under the input-96-predict-48 setting. Notably, for \method, both the background regions and the predicted line segments are highlighted with various colors to explicitly illustrate the distinct dynamic scheduling segments.}
    \label{pems04}
\end{figure*}

\begin{figure*}[htbp]
    \centering
    \includegraphics[width=1\linewidth]{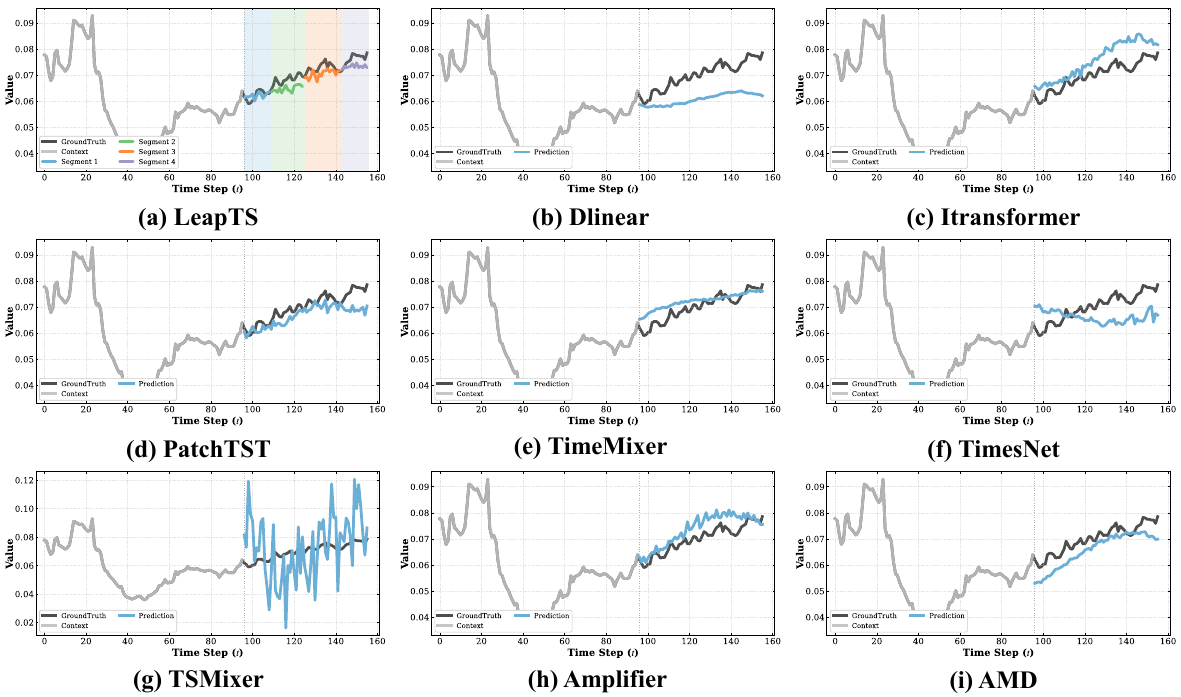}
    \caption{Visualization of forecasting cases from Weather by different models under the input-96-predict-60 setting. Notably, for \method, both the background regions and the predicted line segments are highlighted with various colors to explicitly illustrate the distinct dynamic scheduling segments.}
    \label{weather}
\end{figure*}

\section{Full Results}
\label{appendx:full_results}

In this section, we provide the complete experimental results to supplement the main paper. Table~\ref{tab:full_long_term_forecasting_results} presents the full multivariate forecasting performance across all real-world datasets. Furthermore, Table~\ref{tab:m4_results_univariate} details the full univariate forecasting results on the M4 Competition benchmark \citep{makridakis2018m4} across different frequencies. Furthermore, we include long-term forecasting evaluations on several representative datasets, where the baseline results are directly adopted from Phaseformer \citep{niu2025phaseformer} for comparison (Table \ref{tab:long_forecasting_partial}). The extensive experiments on these widely recognized benchmarks consistently demonstrate the superior effectiveness and robust generalization of \method\ across diverse forecasting scenarios. 

\section{Limitations and Future Work}
\label{appendx:Limitations}

While \method\ achieves strong performance, it still has certain limitations. Currently, our scheduling method  is trained from scratch on individual datasets. Due to its limited model size, it mainly learns patterns within specific domains. To solve this, a promising future direction is to build a large-scale, universal scheduling model. By pre-training on massive and diverse time-series data, we aim to learn general scheduling rules. This would allow the model to perform zero-shot forecasting on various new scenarios without the need for retraining.

\section{Broader Impacts}
\label{appendx:Broader}

The development of \method\ introduces several positive broader impacts for time series community and real-world applications:

\textbf{A Novel Paradigm for Time Series Forecasting.} Our work presents a paradigm shift in time series forecasting from static mappings to dynamic scheduling. By continuously adjusting its forecasting strategies in response to the continuously updating context, the model can inherently adapt to evolving data distributions. This adaptive nature provides a robust foundation for achieving higher-precision predictions in highly non-stationary environments.

\textbf{Potential in Complex Systems.} Beyond standard real-world benchmarks, our framework is particularly well-suited for modeling highly complex systems. In these scientific domains, where the underlying dynamics are often irregular and heavily influenced by unobserved factors, our dynamic scheduling mechanism offers a reliable and powerful tool for high-precision forecasting.

\textbf{Scheduling Traceability.}  \method\ provides an explicit and traceable forecasting process. By revealing exactly how the model dynamically selects temporal scales and step lengths based on data volatility, it helps practitioners observe the internal scheduling dynamics. Such process traceability is especially critical for industrial forecasting, as it enables practitioners to diagnose model behaviors based on the explicitly generated scheduling traces.

\begin{table*}[htbp!]
    \caption{Full multivariate forecasting results on real-world datasets. The input sequence length is set to 36 for the ILI dataset and 96 for all other datasets. \textbf{\textit{AVG}} refers to the overall average across all horizons. The \best{best} and \second{second-best} performances are highlighted.}
    \label{tab:full_long_term_forecasting_results}
  \vskip 0.05in
  \centering
  \resizebox{\textwidth}{!}{
  \begin{threeparttable}
  \begin{small}
  \renewcommand{\multirowsetup}{\centering}
  \renewcommand{\arraystretch}{0.8}
  \setlength{\tabcolsep}{1.5pt}
  \begin{tabular}{c|c|cc|cc|cc|cc|cc|cc|cc|cc|cc}
    \toprule
    \multicolumn{2}{c}{\multirow{2}{*}{Models}} & \multicolumn{2}{c}{\textbf{\method}} &
    \multicolumn{2}{c}{Amplifier} & \multicolumn{2}{c}{AMD} & \multicolumn{2}{c}{TimeMixer} &
    \multicolumn{2}{c}{PatchTST} & \multicolumn{2}{c}{iTransformer} & \multicolumn{2}{c}{TSMixer} &
    \multicolumn{2}{c}{DLinear} & \multicolumn{2}{c}{TimesNet} \\
    \multicolumn{2}{c}{} & \multicolumn{2}{c}{\scalebox{0.8}{\textbf{(Ours)}}} &
    \multicolumn{2}{c}{\scalebox{0.8}{(\citeyear{fei2025amplifier})}} & \multicolumn{2}{c}{\scalebox{0.8}{(\citeyear{hu2025adaptive})}} &
    \multicolumn{2}{c}{\scalebox{0.8}{(\citeyear{wang2024timemixer})}} & \multicolumn{2}{c}{\scalebox{0.8}{(\citeyear{Yuqietal-2023-PatchTST})}}&
    \multicolumn{2}{c}{\scalebox{0.8}{(\citeyear{liu2024itransformer})}}& \multicolumn{2}{c}{\scalebox{0.8}{(\citeyear{chen2023tsmixer})}}&
    \multicolumn{2}{c}{\scalebox{0.8}{(\citeyear{zeng2023transformers})}}& \multicolumn{2}{c}{\scalebox{0.8}{(\citeyear{wu2023timesnet})}} \\
    \cmidrule(lr){3-4} \cmidrule(lr){5-6}\cmidrule(lr){7-8} \cmidrule(lr){9-10}\cmidrule(lr){11-12}\cmidrule(lr){13-14}\cmidrule(lr){15-16}\cmidrule(lr){17-18}\cmidrule(lr){19-20}
    \multicolumn{2}{c}{\textbf{Metric}} & \textbf{MSE} & \textbf{MAE} & \textbf{MSE} & \textbf{MAE} & \textbf{MSE} & \textbf{MAE} & \textbf{MSE} & \textbf{MAE} & \textbf{MSE} & \textbf{MAE} & \textbf{MSE} & \textbf{MAE} & \textbf{MSE} & \textbf{MAE} & \textbf{MSE} & \textbf{MAE} & \textbf{MSE} & \textbf{MAE} \\
    \toprule

    & 18 & \second{0.088} & \best{0.111} & 0.091 & 0.124 & 0.098 & 0.137 & \best{0.087} & \second{0.112} & 0.092 & 0.122 & 0.093 & 0.120 & 0.097 & 0.151 & 0.105 & 0.154 & 0.102 & 0.148 \\
    & 24 & \best{0.097} & \best{0.120} & \best{0.097} & \second{0.128} & 0.110 & 0.152 & \second{0.099} & 0.130 & 0.109 & 0.144 & 0.105 & 0.136 & 0.111 & 0.171 & 0.119 & 0.172 & 0.114 & 0.163 \\
    & 36 & \best{0.114} & \best{0.145} & 0.119 & \second{0.155} & 0.132 & 0.176 & \second{0.118} & 0.156 & 0.127 & 0.168 & 0.125 & 0.160 & 0.133 & 0.199 & 0.143 & 0.201 & 0.125 & 0.175 \\
    & 48 & \best{0.127} & \best{0.162} & \best{0.127} & \second{0.167} & 0.147 & 0.195 & \second{0.131} & 0.174 & 0.144 & 0.186 & 0.136 & 0.173 & 0.149 & 0.219 & 0.161 & 0.224 & 0.138 & 0.190 \\
    & 60 & \second{0.140} & \best{0.178} & \best{0.138} & \second{0.180} & 0.157 & 0.206 & \second{0.140} & 0.184 & 0.153 & 0.198 & 0.147 & 0.186 & 0.160 & 0.233 & 0.172 & 0.232 & 0.145 & 0.198 \\
    \rowcolor{rowgray} \cellcolor{white}\multirow{-6}{*}{\rotatebox{90}{\textbf{Weather}}}
    & \textit{\textbf{AVG}} & \best{0.113} & \best{0.143} & \second{0.114} & \second{0.151} & 0.129 & 0.173 & 0.115 & \second{0.151} & 0.125 & 0.164 & 0.121 & 0.155 & 0.130 & 0.195 & 0.140 & 0.197 & 0.125 & 0.175 \\
    \midrule

    & 18 & \best{0.103} & \second{0.199} & \best{0.103} & 0.200 & 0.126 & 0.223 & \second{0.114} & 0.209 & 0.135 & 0.235 & \best{0.103} & \best{0.198} & 0.137 & 0.253 & 0.175 & 0.273 & 0.132 & 0.241 \\
    & 24 & \second{0.108} & \best{0.203} & \best{0.106} & \second{0.204} & 0.133 & 0.228 & 0.120 & 0.215 & 0.144 & 0.242 & 0.110 & \second{0.204} & 0.146 & 0.261 & 0.181 & 0.276 & 0.134 & 0.242 \\
    & 36 & \best{0.119} & \best{0.215} & 0.125 & 0.222 & 0.148 & 0.244 & 0.133 & 0.227 & 0.161 & 0.256 & \second{0.123} & \second{0.216} & 0.162 & 0.275 & 0.199 & 0.290 & 0.144 & 0.250 \\
    & 48 & \best{0.126} & \best{0.222} & 0.134 & 0.231 & 0.156 & 0.249 & 0.142 & 0.236 & 0.172 & 0.265 & \second{0.133} & \second{0.225} & 0.181 & 0.288 & 0.210 & 0.298 & 0.149 & 0.254 \\
    & 60 & \best{0.131} & \best{0.226} & 0.142 & 0.239 & 0.163 & 0.255 & 0.148 & 0.241 & 0.178 & 0.267 & \second{0.141} & \second{0.232} & 0.201 & 0.304 & 0.215 & 0.302 & 0.157 & 0.261 \\
    \rowcolor{rowgray} \cellcolor{white}\multirow{-6}{*}{\rotatebox{90}{\textbf{Electricity}}}
    & \textit{\textbf{AVG}} & \best{0.117} & \best{0.213} & \second{0.122} & 0.219 & 0.145 & 0.240 & 0.131 & 0.226 & 0.158 & 0.253 & \second{0.122} & \second{0.215} & 0.165 & 0.276 & 0.196 & 0.288 & 0.143 & 0.250 \\
    \midrule

    & 12 & \best{1.040} & \best{0.609} & 1.588 & 0.754 & 1.865 & 0.844 & 1.563 & 0.763 & 1.766 & 0.805 & 1.702 & 0.862 & 2.246 & 0.994 & 4.656 & 1.614 & \second{1.341} & \second{0.736} \\
    & 18 & \second{1.374} & \best{0.693} & 1.661 & 0.807 & 2.197 & 0.932 & 1.955 & 0.842 & 2.464 & 0.922 & \best{1.348} & \second{0.729} & 2.706 & 1.116 & 4.807 & 1.659 & 1.601 & 0.811 \\
    & 24 & 1.861 & \best{0.813} & \second{1.792} & \second{0.831} & 2.466 & 0.984 & 2.661 & 0.986 & 2.370 & 0.912 & 1.985 & 0.893 & 3.045 & 1.200 & 5.061 & 1.710 & \best{1.637} & 0.866 \\
    & 36 & \second{1.948} & \second{0.837} & \best{1.703} & \best{0.832} & 2.465 & 0.987 & 2.305 & 0.941 & 2.063 & 0.874 & 2.130 & 0.936 & 3.384 & 1.270 & 4.414 & 1.550 & 2.678 & 0.987 \\
    \rowcolor{rowgray} \cellcolor{white}\multirow{-5}{*}{\rotatebox{90}{\textbf{ILI}}}
    & \textit{\textbf{AVG}} & \best{1.556} & \best{0.738} & \second{1.686} & \second{0.806} & 2.248 & 0.937 & 2.121 & 0.883 & 2.166 & 0.878 & 1.791 & 0.855 & 2.845 & 1.145 & 4.735 & 1.633 & 1.814 & 0.850 \\
    \midrule

    & 18 & \best{0.282} & \best{0.334} & 0.290 & 0.344 & 0.294 & 0.345 & \second{0.287} & \second{0.343} & 0.290 & 0.347 & 0.304 & 0.355 & 0.352 & 0.405 & 0.324 & 0.370 & 0.336 & 0.382 \\
    & 24 & \best{0.294} & \best{0.341} & \second{0.297} & 0.350 & 0.300 & \second{0.348} & 0.303 & 0.355 & 0.298 & 0.352 & 0.313 & 0.361 & 0.367 & 0.417 & 0.328 & 0.371 & 0.339 & 0.387 \\
    & 36 & \best{0.319} & \best{0.355} & 0.328 & 0.366 & 0.324 & \second{0.361} & \second{0.323} & 0.366 & \second{0.323} & 0.368 & 0.337 & 0.376 & 0.403 & 0.442 & 0.348 & 0.383 & 0.366 & 0.400 \\
    & 48 & \best{0.333} & \best{0.364} & 0.339 & 0.372 & \second{0.334} & \second{0.367} & 0.336 & 0.376 & 0.338 & 0.377 & 0.349 & 0.383 & 0.418 & 0.452 & 0.358 & 0.389 & 0.364 & 0.399 \\
    & 60 & \best{0.351} & \best{0.374} & \second{0.355} & \second{0.378} & 0.356 & 0.379 & 0.367 & 0.387 & \second{0.355} & 0.385 & 0.364 & 0.392 & 0.443 & 0.468 & 0.371 & 0.397 & 0.371 & 0.403 \\
    \rowcolor{rowgray} \cellcolor{white}\multirow{-6}{*}{\rotatebox{90}{\textbf{ETTh1}}}
    & \textit{\textbf{AVG}} & \best{0.316} & \best{0.354} & 0.322 & 0.362 & 0.322 & \second{0.360} & 0.323 & 0.365 & \second{0.321} & 0.366 & 0.333 & 0.373 & 0.397 & 0.437 & 0.346 & 0.382 & 0.355 & 0.394 \\
    \midrule

    & 18 & \best{0.151} & \best{0.244} & 0.163 & 0.256 & \second{0.155} & \second{0.252} & \second{0.155} & 0.253 & 0.157 & 0.268 & 0.164 & 0.261 & 0.222 & 0.348 & 0.176 & 0.279 & 0.179 & 0.277 \\
    & 24 & \best{0.166} & \best{0.254} & 0.178 & 0.269 & \second{0.170} & \second{0.262} & 0.175 & 0.267 & 0.178 & 0.268 & 0.183 & 0.275 & 0.265 & 0.385 & 0.193 & 0.291 & 0.199 & 0.289 \\
    & 36 & \best{0.199} & \best{0.278} & 0.206 & 0.288 & \best{0.199} & \second{0.281} & \second{0.205} & 0.285 & 0.213 & 0.290 & 0.217 & 0.297 & 0.361 & 0.463 & 0.228 & 0.317 & 0.224 & 0.304 \\
    & 48 & \second{0.223} & \best{0.292} & 0.231 & 0.302 & \best{0.221} & \second{0.295} & \second{0.223} & 0.296 & 0.235 & 0.303 & 0.241 & 0.312 & 0.495 & 0.546 & 0.257 & 0.339 & 0.252 & 0.320 \\
    & 60 & \second{0.246} & \best{0.309} & 0.252 & 0.317 & \best{0.240} & \best{0.309} & 0.249 & \second{0.314} & 0.260 & 0.325 & 0.260 & 0.324 & 0.647 & 0.633 & 0.283 & 0.357 & 0.267 & 0.333 \\
    \rowcolor{rowgray} \cellcolor{white}\multirow{-6}{*}{\rotatebox{90}{\textbf{ETTh2}}}
    & \textit{\textbf{AVG}} & \best{0.197} & \best{0.275} & 0.206 & 0.286 & \best{0.197} & \second{0.280} & \second{0.201} & 0.283 & 0.209 & 0.291 & 0.213 & 0.294 & 0.398 & 0.475 & 0.227 & 0.317 & 0.224 & 0.305 \\
    \midrule

    & 18 & \best{0.177} & \best{0.257} & \second{0.180} & \second{0.265} & 0.194 & 0.278 & 0.183 & 0.267 & 0.181 & 0.267 & 0.192 & 0.274 & 0.214 & 0.301 & 0.215 & 0.289 & 0.195 & 0.278 \\
    & 24 & \second{0.210} & \best{0.281} & 0.212 & \second{0.285} & 0.227 & 0.300 & \best{0.207} & \best{0.281} & 0.217 & 0.291 & 0.230 & 0.301 & 0.253 & 0.327 & 0.251 & 0.313 & 0.238 & 0.307 \\
    & 36 & \best{0.255} & \best{0.311} & \second{0.260} & \second{0.320} & 0.269 & 0.327 & 0.266 & 0.321 & 0.265 & 0.324 & 0.278 & 0.333 & 0.324 & 0.372 & 0.294 & 0.341 & 0.285 & 0.338 \\
    & 48 & \best{0.282} & \best{0.330} & \second{0.286} & 0.338 & 0.292 & 0.341 & 0.287 & \second{0.336} & 0.288 & 0.340 & 0.322 & 0.363 & 0.371 & 0.405 & 0.317 & 0.354 & 0.303 & 0.350 \\
    & 60 & \second{0.298} & \best{0.339} & 0.299 & 0.345 & 0.306 & 0.349 & \best{0.295} & \second{0.343} & 0.305 & 0.351 & 0.336 & 0.373 & 0.401 & 0.421 & 0.329 & 0.362 & 0.327 & 0.365 \\
    \rowcolor{rowgray} \cellcolor{white}\multirow{-6}{*}{\rotatebox{90}{\textbf{ETTm1}}}
    & \textit{\textbf{AVG}} & \best{0.244} & \best{0.304} & \second{0.247} & 0.311 & 0.258 & 0.319 & 0.248 & \second{0.310} & 0.251 & 0.315 & 0.272 & 0.329 & 0.313 & 0.365 & 0.281 & 0.332 & 0.270 & 0.328 \\
    \midrule

    & 18 & \best{0.088} & \best{0.180} & 0.091 & \second{0.182} & 0.098 & 0.197 & \second{0.089} & 0.183 & 0.090 & 0.183 & 0.092 & 0.186 & 0.118 & 0.238 & 0.096 & 0.198 & 0.098 & 0.195 \\
    & 24 & \second{0.100} & \second{0.193} & 0.102 & 0.194 & 0.110 & 0.210 & \best{0.098} & \best{0.192} & 0.101 & 0.197 & 0.105 & 0.201 & 0.120 & 0.235 & 0.109 & 0.214 & 0.104 & 0.201 \\
    & 36 & \best{0.119} & \best{0.212} & 0.125 & 0.220 & 0.129 & 0.229 & \best{0.119} & \second{0.216} & \second{0.120} & \second{0.216} & 0.123 & 0.219 & 0.160 & 0.283 & 0.131 & 0.239 & 0.127 & 0.222 \\
    & 48 & \best{0.134} & \best{0.225} & 0.136 & \second{0.227} & 0.144 & 0.242 & \second{0.135} & 0.228 & 0.136 & 0.231 & 0.139 & 0.234 & 0.187 & 0.316 & 0.148 & 0.256 & 0.142 & 0.234 \\
    & 60 & \best{0.147} & \best{0.235} & \second{0.148} & \second{0.238} & 0.156 & 0.250 & \second{0.148} & 0.240 & 0.153 & 0.244 & 0.155 & 0.247 & 0.224 & 0.352 & 0.161 & 0.266 & 0.160 & 0.246 \\
    \rowcolor{rowgray} \cellcolor{white}\multirow{-6}{*}{\rotatebox{90}{\textbf{ETTm2}}}
    & \textit{\textbf{AVG}} & \best{0.118} & \best{0.209} & \second{0.120} & \second{0.212} & 0.127 & 0.226 & \best{0.118} & \second{0.212} & \second{0.120} & 0.214 & 0.123 & 0.217 & 0.162 & 0.285 & 0.129 & 0.235 & 0.126 & 0.220 \\
    \midrule

    & 18 & \second{0.020} & \best{0.091} & 0.022 & 0.096 & 0.021 & 0.097 & \second{0.020} & \second{0.093} & \best{0.019} & \best{0.091} & 0.022 & 0.099 & 0.070 & 0.205 & 0.030 & 0.126 & 0.027 & 0.116 \\
    & 24 & \second{0.025} & \second{0.105} & \second{0.025} & 0.107 & 0.026 & 0.109 & 0.026 & 0.108 & \best{0.024} & \best{0.103} & 0.028 & 0.116 & 0.074 & 0.212 & 0.034 & 0.134 & 0.032 & 0.128 \\
    & 36 & \best{0.034} & \best{0.126} & 0.042 & 0.138 & \second{0.035} & \second{0.128} & 0.037 & 0.132 & 0.036 & 0.131 & 0.039 & 0.137 & 0.081 & 0.220 & 0.043 & 0.151 & 0.045 & 0.153 \\
    & 48 & \best{0.044} & \best{0.144} & 0.049 & 0.152 & \best{0.044} & \second{0.145} & 0.047 & 0.150 & \second{0.045} & \second{0.145} & 0.046 & 0.151 & 0.087 & 0.227 & 0.053 & 0.169 & 0.059 & 0.173 \\
    & 60 & \best{0.053} & \best{0.159} & \second{0.055} & 0.165 & \best{0.053} & \second{0.160} & 0.056 & 0.164 & \second{0.055} & 0.161 & 0.065 & 0.179 & 0.215 & 0.376 & 0.062 & 0.182 & 0.073 & 0.197 \\
    \rowcolor{rowgray} \cellcolor{white}\multirow{-6}{*}{\rotatebox{90}{\textbf{Exchange}}}
    & \textit{\textbf{AVG}} & \best{0.035} & \best{0.125} & 0.039 & 0.132 & \second{0.036} & 0.128 & 0.037 & 0.129 & \second{0.036} & \second{0.126} & 0.040 & 0.136 & 0.105 & 0.248 & 0.044 & 0.152 & 0.047 & 0.153 \\
    \midrule

    & 18 & \second{0.096} & \best{0.199} & 0.137 & 0.255 & 0.147 & 0.262 & 0.099 & 0.211 & 0.143 & 0.265 & 0.103 & 0.209 & \best{0.095} & 0.214 & 0.173 & 0.295 & \second{0.096} & \second{0.206} \\
    & 24 & \second{0.110} & \best{0.213} & 0.156 & 0.271 & 0.187 & 0.298 & 0.120 & 0.234 & 0.176 & 0.290 & 0.120 & 0.228 & \best{0.105} & 0.226 & 0.210 & 0.327 & \best{0.105} & \second{0.216} \\
    & 36 & 0.140 & \second{0.239} & 0.216 & 0.321 & 0.266 & 0.357 & 0.152 & 0.263 & 0.254 & 0.354 & 0.153 & 0.259 & \best{0.122} & 0.246 & 0.281 & 0.385 & \second{0.123} & \best{0.235} \\
    & 48 & 0.171 & 0.264 & 0.240 & 0.345 & 0.348 & 0.414 & 0.182 & 0.288 & 0.337 & 0.416 & 0.185 & 0.288 & \best{0.135} & \second{0.262} & 0.337 & 0.425 & \second{0.143} & \best{0.257} \\
    \rowcolor{rowgray} \cellcolor{white}\multirow{-5}{*}{\rotatebox{90}{\textbf{PEMS04}}}
    & \textit{\textbf{AVG}} & 0.129 & \best{0.229} & 0.187 & 0.298 & 0.237 & 0.333 & 0.138 & 0.249 & 0.227 & 0.331 & 0.140 & 0.246 & \best{0.114} & \second{0.237} & 0.250 & 0.358 & \second{0.117} & \best{0.229} \\
    \midrule

    & 18 & \best{0.079} & \best{0.176} & 0.122 & 0.223 & 0.117 & 0.233 & 0.086 & 0.194 & 0.114 & 0.234 & \second{0.080} & \second{0.181} & 0.083 & 0.199 & 0.154 & 0.277 & 0.090 & 0.193 \\
    & 24 & \second{0.094} & \best{0.191} & 0.139 & 0.255 & 0.161 & 0.284 & 0.109 & 0.224 & 0.144 & 0.259 & \best{0.093} & \second{0.196} & \best{0.093} & 0.211 & 0.202 & 0.320 & 0.099 & 0.204 \\
    & 36 & 0.124 & \best{0.216} & 0.207 & 0.319 & 0.231 & 0.338 & 0.135 & 0.242 & 0.211 & 0.317 & 0.120 & 0.225 & \best{0.110} & 0.230 & 0.299 & 0.393 & \second{0.114} & \second{0.220} \\
    & 48 & 0.157 & \second{0.244} & 0.247 & 0.350 & 0.298 & 0.382 & 0.178 & 0.280 & 0.279 & 0.366 & 0.145 & 0.249 & \best{0.127} & 0.245 & 0.385 & 0.449 & \second{0.132} & \best{0.239} \\
    \rowcolor{rowgray} \cellcolor{white}\multirow{-5}{*}{\rotatebox{90}{\textbf{PEMS07}}}
    & \textit{\textbf{AVG}} & 0.114 & \best{0.207} & 0.178 & 0.287 & 0.202 & 0.309 & 0.127 & 0.235 & 0.187 & 0.294 & \second{0.109} & \second{0.213} & \best{0.103} & 0.221 & 0.260 & 0.360 & \second{0.109} & 0.214 \\
    \midrule

    & 12 & 2.921 & \best{0.943} & \best{2.876} & \second{0.996} & 3.911 & 1.099 & 7.423 & 1.781 & 3.245 & 1.065 & \second{2.887} & 1.004 & 8.030 & 1.785 & 9.809 & 1.987 & 7.800 & 1.743 \\
    & 24 & \second{5.221} & \best{1.404} & 5.728 & 1.530 & 6.109 & 1.509 & 12.612 & 2.243 & 5.390 & \second{1.420} & \best{5.220} & 1.456 & 9.833 & 1.936 & 10.510 & 2.015 & 10.562 & 2.001 \\
    \rowcolor{rowgray} \cellcolor{white}\multirow{-3.2}{*}{\raisebox{-1.5ex}{\rotatebox{90}{\textbf{COVID}}}}
    & \textit{\textbf{AVG}} & \second{4.071} & \best{1.174} & 4.302 & 1.263 & 5.010 & 1.304 & 10.017 & 2.012 & 4.317 & 1.243 & \best{4.054} & \second{1.230} & 8.932 & 1.860 & 10.160 & 2.001 & 9.181 & 1.872 \\
    \midrule

    & 12 & \best{0.094} & \best{0.218} & \second{0.105} & \second{0.234} & 0.142 & 0.276 & 0.113 & 0.244 & 0.121 & 0.258 & 0.150 & 0.285 & 0.182 & 0.323 & 0.201 & 0.353 & 0.109 & 0.236 \\
    & 24 & \best{0.161} & \best{0.285} & 0.190 & 0.312 & 0.176 & 0.306 & \second{0.171} & \second{0.297} & 0.187 & 0.319 & 0.234 & 0.358 & 0.268 & 0.396 & 0.268 & 0.407 & 0.180 & 0.300 \\
    & 36 & \best{0.218} & \best{0.332} & 0.252 & 0.373 & \second{0.223} & \second{0.341} & 0.234 & 0.357 & 0.260 & 0.380 & 0.288 & 0.414 & 0.338 & 0.443 & 0.336 & 0.460 & 0.248 & 0.357 \\
    \rowcolor{rowgray} \cellcolor{white}\multirow{-4}{*}{\rotatebox{90}{\textbf{SP500}}}
    & \textit{\textbf{AVG}} & \best{0.158} & \best{0.278} & 0.182 & 0.306 & 0.181 & 0.307 & \second{0.172} & 0.299 & 0.189 & 0.319 & 0.224 & 0.352 & 0.263 & 0.387 & 0.268 & 0.407 & 0.179 & \second{0.298} \\
    \midrule

    & 12 & \best{6.797} & \best{0.418} & 6.879 & 0.458 & 7.248 & 0.464 & 7.051 & 0.497 & 6.811 & 0.430 & \second{6.798} & \second{0.428} & 7.983 & 0.810 & 7.013 & 0.512 & 7.307 & 0.545 \\
    & 24 & 6.887 & \best{0.442} & 6.974 & 0.492 & 6.942 & 0.470 & 6.974 & 0.485 & \second{6.881} & \second{0.459} & \best{6.870} & \second{0.459} & 8.049 & 0.839 & 7.077 & 0.550 & 7.298 & 0.554 \\
    & 36 & \second{6.421} & \best{0.458} & 6.708 & 0.564 & 6.467 & \second{0.481} & 6.711 & 0.548 & 6.487 & 0.492 & \best{6.391} & \best{0.458} & 7.644 & 0.865 & 6.624 & 0.579 & 6.923 & 0.607 \\
    \rowcolor{rowgray} \cellcolor{white}\multirow{-4}{*}{\rotatebox{90}{\textbf{Wiki}}}
    & \textit{\textbf{AVG}} & \second{6.702} & \best{0.440} & 6.854 & 0.505 & 6.886 & 0.472 & 6.912 & 0.510 & 6.727 & 0.461 & \best{6.687} & \second{0.448} & 7.892 & 0.838 & 6.905 & 0.547 & 7.176 & 0.569 \\
    \midrule

    \rowcolor{rowblue} \multicolumn{2}{c|}{\textit{\textbf{Average}}}
    & \best{1.067} & \best{0.361} & 1.120 & 0.395 & 1.229 & 0.414 & 1.589 & 0.451 & 1.156 & 0.404 & \second{1.095} & \second{0.390} & 1.678 & 0.536 & 1.842 & 0.554 & 1.528 & 0.450 \\
    \midrule
    \rowcolor{rowpink} \multicolumn{2}{c|}{\textit{\textbf{1\textsuperscript{st} Count}}}
    & \best{31} & \best{48} & \second{7} & 1 & 5 & 1 & 5 & 2 & 2 & 2 & 6 & 2 & \second{7} & 0 & 0 & 0 & 2 & \second{3} \\
    \bottomrule
  \end{tabular}
  \end{small}
  \end{threeparttable}
  }
\end{table*}

\newpage

\begin{table*}[htbp!]
\caption{Full univariate forecasting results on the M4 Competition dataset \citep{makridakis2018m4} across different frequencies. 
\textbf{\textit{Average}} denotes the weighted average across frequency groups following the standard M4 evaluation. 
The \best{best} and \second{second-best} performances are highlighted.}
\label{tab:m4_results_univariate}
\vskip 0.05in
\centering
\resizebox{\textwidth}{!}{
\begin{threeparttable}
\normalsize
\renewcommand{\arraystretch}{1.4}
\setlength{\tabcolsep}{1.15pt}

\newcommand{\num}[1]{{\large #1}}
\newcommand{\met}[1]{{\textbf{#1}}}
\newcommand{\grpbox}[1]{\makebox[6.2em][c]{#1}}
\newcommand{\grp}[1]{{\grpbox{\textit{\textbf{#1}}}}}
\newcommand{\modelhead}[1]{\scalebox{1.03}{#1}}
\newcommand{\yearhead}[1]{\scalebox{0.98}{#1}}

\begin{tabular}{@{}c|c|cccccccccccccccc@{}}
\toprule
\multirow{2}{*}{\textbf{Dataset}} & \multirow{2}{*}{\textbf{Metric}} &
\multicolumn{1}{c}{\modelhead{\textbf{\method}}} &
\multicolumn{1}{c}{\modelhead{\textbf{Amplifier}}} &
\multicolumn{1}{c}{\modelhead{\textbf{TimeMixer}}} &
\multicolumn{1}{c}{\modelhead{\textbf{iTrans.}}} &
\multicolumn{1}{c}{\modelhead{\textbf{TiDE}}} &
\multicolumn{1}{c}{\modelhead{\textbf{TNet}}} &
\multicolumn{1}{c}{\modelhead{\textbf{N-HiTS}}} &
\multicolumn{1}{c}{\modelhead{\textbf{DLinear}}} &
\multicolumn{1}{c}{\modelhead{\textbf{Patch}}} &
\multicolumn{1}{c}{\modelhead{\textbf{MICN}}} &
\multicolumn{1}{c}{\modelhead{\textbf{FiLM}}} &
\multicolumn{1}{c}{\modelhead{\textbf{LightTS}}} &
\multicolumn{1}{c}{\modelhead{\textbf{FED}}} &
\multicolumn{1}{c}{\modelhead{\textbf{Stat.}}} &
\multicolumn{1}{c}{\modelhead{\textbf{Auto}}} &
\multicolumn{1}{c}{\modelhead{\textbf{N-BEATS}}} \\

& &
\multicolumn{1}{c}{\yearhead{(Ours)}} &
\multicolumn{1}{c}{\yearhead{(\citeyear{fei2025amplifier})}} &
\multicolumn{1}{c}{\yearhead{(\citeyear{wang2024timemixer})}} &
\multicolumn{1}{c}{\yearhead{(\citeyear{liu2024itransformer})}} &
\multicolumn{1}{c}{\yearhead{(\citeyear{das2023longterm})}} &
\multicolumn{1}{c}{\yearhead{(\citeyear{wu2023timesnet})}} &
\multicolumn{1}{c}{\yearhead{(\citeyear{challu2023nhits})}} &
\multicolumn{1}{c}{\yearhead{(\citeyear{zeng2023transformers})}} &
\multicolumn{1}{c}{\yearhead{(\citeyear{Yuqietal-2023-PatchTST})}} &
\multicolumn{1}{c}{\yearhead{(\citeyear{wang2023micn})}} &
\multicolumn{1}{c}{\yearhead{(\citeyear{zhou2022film})}} &
\multicolumn{1}{c}{\yearhead{(\citeyear{campos2023lightts})}} &
\multicolumn{1}{c}{\yearhead{(\citeyear{zhou2022fedformer})}} &
\multicolumn{1}{c}{\yearhead{(\citeyear{liu2022non})}} &
\multicolumn{1}{c}{\yearhead{(\citeyear{wu2021autoformer})}} &
\multicolumn{1}{c}{\yearhead{(\citeyear{Oreshkin2020N-BEATS})}} \\
\midrule

\multirow{3}{*}{\grp{\large Yearly}}
& \met{SMAPE} & \best{\num{13.182}} & \num{13.372} & \second{\num{13.206}} & \num{13.923} & \num{15.320} & \num{13.387} & \num{13.418} & \num{16.965} & \num{16.463} & \num{25.022} & \num{17.431} & \num{14.247} & \num{13.728} & \num{13.717} & \num{13.974} & \num{13.436} \\
& \met{MASE}  & \second{\num{2.956}} & \num{2.977} & \best{\num{2.916}} & \num{3.214} & \num{3.540} & \num{2.996} & \num{3.045} & \num{4.283} & \num{3.967} & \num{7.162} & \num{4.043} & \num{3.109} & \num{3.048} & \num{3.078} & \num{3.134} & \num{3.043} \\
& \met{OWA}   & \best{\num{0.775}} & \num{0.784} & \second{\num{0.776}} & \num{0.830} & \num{0.910} & \num{0.786} & \num{0.793} & \num{1.058} & \num{1.003} & \num{1.667} & \num{1.042} & \num{0.827} & \num{0.803} & \num{0.807} & \num{0.822} & \num{0.794} \\
\midrule

\multirow{3}{*}{\grp{\large Quarterly}}
& \met{SMAPE} & \best{\num{9.979}} & \num{10.227} & \second{\num{9.996}} & \num{10.757} & \num{11.830} & \num{10.100} & \num{10.202} & \num{12.145} & \num{10.644} & \num{15.214} & \num{12.925} & \num{11.364} & \num{10.792} & \num{10.958} & \num{11.338} & \num{10.124} \\
& \met{MASE}  & \best{\num{1.159}} & \num{1.191} & \second{\num{1.166}} & \num{1.283} & \num{1.410} & \num{1.182} & \num{1.194} & \num{1.520} & \num{1.278} & \num{1.963} & \num{1.664} & \num{1.328} & \num{1.283} & \num{1.325} & \num{1.365} & \num{1.169} \\
& \met{OWA}   & \second{\num{0.876}} & \num{0.899} & \best{\num{0.825}} & \num{0.956} & \num{1.050} & \num{0.890} & \num{0.899} & \num{1.106} & \num{0.949} & \num{1.407} & \num{1.193} & \num{1.000} & \num{0.958} & \num{0.981} & \num{1.012} & \num{0.886} \\
\midrule

\multirow{3}{*}{\grp{\large Monthly}}
& \met{SMAPE} & \best{\num{12.540}} & \num{12.969} & \second{\num{12.605}} & \num{13.796} & \num{15.180} & \num{12.670} & \num{12.791} & \num{13.514} & \num{13.399} & \num{16.943} & \num{15.407} & \num{14.014} & \num{14.260} & \num{13.917} & \num{13.958} & \num{12.677} \\
& \met{MASE}  & \best{\num{0.913}} & \num{0.968} & \second{\num{0.919}} & \num{1.083} & \num{1.190} & \num{0.933} & \num{0.969} & \num{1.037} & \num{1.031} & \num{1.442} & \num{1.298} & \num{1.053} & \num{1.102} & \num{1.097} & \num{1.103} & \num{0.937} \\
& \met{OWA}   & \best{\num{0.864}} & \num{0.905} & \second{\num{0.869}} & \num{0.987} & \num{1.090} & \num{0.878} & \num{0.899} & \num{0.956} & \num{0.949} & \num{1.265} & \num{1.144} & \num{0.981} & \num{1.012} & \num{0.998} & \num{1.002} & \num{0.880} \\
\midrule

\multirow{3}{*}{\grp{\large Others}}
& \met{SMAPE} & \best{\num{4.490}} & \num{5.531} & \second{\num{4.564}} & \num{5.569} & \num{6.120} & \num{4.891} & \num{5.061} & \num{6.709} & \num{6.558} & \num{41.985} & \num{7.134} & \num{15.880} & \num{4.954} & \num{6.302} & \num{5.485} & \num{4.925} \\
& \met{MASE}  & \best{\num{2.988}} & \num{3.649} & \second{\num{3.115}} & \num{3.940} & \num{4.330} & \num{3.302} & \num{3.216} & \num{4.953} & \num{4.511} & \num{62.734} & \num{5.090} & \num{11.434} & \num{3.264} & \num{4.064} & \num{3.865} & \num{3.391} \\
& \met{OWA}   & \best{\num{0.944}} & \num{1.135} & \second{\num{0.982}} & \num{1.207} & \num{1.330} & \num{1.035} & \num{1.040} & \num{1.487} & \num{1.401} & \num{14.313} & \num{1.553} & \num{3.474} & \num{1.036} & \num{1.304} & \num{1.187} & \num{1.053} \\
\midrule

\rowcolor{rowblue}
& \met{SMAPE} & \best{\num{11.671}} & \num{12.030} & \second{\num{11.723}} & \num{12.684} & \num{13.950} & \num{11.829} & \num{11.927} & \num{13.639} & \num{13.152} & \num{19.638} & \num{14.863} & \num{13.525} & \num{12.840} & \num{12.780} & \num{12.909} & \num{11.851} \\
\rowcolor{rowblue}
& \met{MASE}  & \best{\num{1.546}} & \num{1.617} & \second{\num{1.559}} & \num{1.764} & \num{1.940} & \num{1.585} & \num{1.613} & \num{2.095} & \num{1.945} & \num{5.947} & \num{2.207} & \num{2.111} & \num{1.701} & \num{1.756} & \num{1.771} & \second{\num{1.559}} \\
\rowcolor{rowblue}
\multirow{-3}{*}{\grp{\large Average}} & \met{OWA} & \best{\num{0.834}} & \num{0.887} & \second{\num{0.840}} & \num{0.929} & \num{1.020} & \num{0.851} & \num{0.861} & \num{1.051} & \num{0.998} & \num{2.279} & \num{1.125} & \num{1.051} & \num{0.918} & \num{0.930} & \num{0.939} & \num{0.855} \\
\bottomrule
\end{tabular}
\end{threeparttable}
}
\end{table*}

\begin{table*}[htbp!]
\caption{Long-term forecasting results on representative datasets. The input sequence length is set to 720 for all models. The \best{best} and \second{second-best} results are highlighted.}  \label{tab:long_forecasting_partial}
  \vskip 0.05in
  \centering
  \resizebox{\textwidth}{!}{
  \begin{threeparttable}
  \begin{small}
  \renewcommand{\multirowsetup}{\centering}
  \renewcommand{\arraystretch}{1.0}
  \setlength{\tabcolsep}{1.5pt}
  \begin{tabular}{cc|cc|cc|cc|cc|cc|cc|cc|cc|cc|cc}
    \toprule
    \multicolumn{2}{c|}{\textbf{Models}} & \multicolumn{2}{c}{\textbf{LeapTS}} & \multicolumn{2}{c}{Phaseformer} & \multicolumn{2}{c}{PatchTST} & \multicolumn{2}{c}{iTransformer} & \multicolumn{2}{c}{Crossformer} & \multicolumn{2}{c}{FEDformer} & \multicolumn{2}{c}{TimeBase} & \multicolumn{2}{c}{SparseTSF} & \multicolumn{2}{c}{FITS} & \multicolumn{2}{c}{TimeMixer} \\
    \cmidrule(lr){3-4} \cmidrule(lr){5-6} \cmidrule(lr){7-8} \cmidrule(lr){9-10} \cmidrule(lr){11-12} \cmidrule(lr){13-14} \cmidrule(lr){15-16} \cmidrule(lr){17-18} \cmidrule(lr){19-20} \cmidrule(lr){21-22}
    \multicolumn{2}{c|}{\textbf{Metric}} & \textbf{MSE} & \textbf{MAE} & \textbf{MSE} & \textbf{MAE} & \textbf{MSE} & \textbf{MAE} & \textbf{MSE} & \textbf{MAE} & \textbf{MSE} & \textbf{MAE} & \textbf{MSE} & \textbf{MAE} & \textbf{MSE} & \textbf{MAE} & \textbf{MSE} & \textbf{MAE} & \textbf{MSE} & \textbf{MAE} & \textbf{MSE} & \textbf{MAE} \\
    \toprule

    & 96  & \best{0.292} & \best{0.342} & \second{0.293} & \second{0.344} & 0.298 & 0.352 & 0.315 & 0.369 & 0.306 & 0.353 & 0.406 & 0.441 & 0.311 & 0.351 & 0.314 & 0.359 & 0.313 & 0.357 & 0.332 & 0.384 \\
    & 192 & \best{0.323} & \best{0.360} & \best{0.323} & \second{0.361} & \second{0.335} & 0.373 & 0.349 & 0.388 & 0.341 & 0.385 & 0.450 & 0.477 & 0.338 & 0.371 & 0.348 & 0.376 & 0.339 & 0.369 & 0.362 & 0.398 \\
    & 336 & \second{0.362} & \best{0.381} & \best{0.358} & \best{0.381} & 0.366 & 0.389 & 0.381 & 0.409 & 0.383 & 0.420 & 0.436 & 0.466 & 0.364 & 0.386 & 0.368 & 0.386 & 0.367 & \second{0.385} & 0.386 & 0.413 \\
    \multirow{-4}{*}{\rotatebox{90}{\textbf{ETTm1}}} 
    & 720 & 0.417 & 0.419 & \best{0.412} & \best{0.410} & 0.420 & 0.421 & 0.437 & 0.439 & 0.532 & 0.512 & 0.462 & 0.479 & \second{0.413} & 0.414 & 0.419 & \second{0.413} & 0.417 & 0.417 & 0.452 & 0.457 \\
    \midrule

    & 96  & \best{0.159} & \best{0.252} & 0.163 & \second{0.256} & 0.165 & 0.260 & 0.179 & 0.274 & 0.244 & 0.338 & 0.339 & 0.406 & \second{0.162} & \second{0.256} & 0.167 & 0.259 & 0.166 & \second{0.256} & 0.192 & 0.285 \\
    & 192 & \best{0.216} & \best{0.290} & 0.219 & \second{0.293} & 0.219 & 0.298 & 0.239 & 0.314 & 0.350 & 0.412 & 0.397 & 0.452 & \second{0.218} & \second{0.293} & 0.219 & 0.297 & 0.271 & 0.328 & 0.307 & 0.362 \\
    & 336 & \best{0.268} & \best{0.326} & \second{0.269} & \best{0.326} & \best{0.268} & 0.333 & 0.309 & 0.356 & 0.400 & 0.431 & 0.418 & 0.452 & 0.270 & \second{0.328} & 0.271 & 0.330 & 0.352 & 0.380 & 0.380 & 0.412 \\
    \multirow{-4}{*}{\rotatebox{90}{\textbf{ETTm2}}} 
    & 720 & \best{0.351} & 0.383 & \best{0.351} & \best{0.379} & \second{0.352} & 0.386 & 0.387 & 0.407 & 0.574 & 0.525 & 0.451 & 0.499 & \second{0.352} & \second{0.380} & 0.353 & \second{0.380} & \second{0.352} & \second{0.380} & 0.380 & 0.412 \\
    \midrule

    & 96  & \best{0.146} & \second{0.197} & \second{0.148} & \best{0.195} & 0.149 & 0.199 & 0.159 & 0.212 & 0.151 & 0.210 & 0.289 & 0.342 & \best{0.146} & 0.198 & 0.174 & 0.231 & 0.176 & 0.232 & 0.163 & 0.223 \\
    & 192 & \second{0.189} & \best{0.236} & 0.193 & \second{0.237} & 0.193 & 0.243 & 0.203 & 0.252 & 0.220 & 0.273 & 0.340 & 0.403 & \best{0.185} & 0.241 & 0.216 & 0.267 & 0.203 & 0.256 & 0.201 & 0.255 \\
    & 336 & \second{0.242} & \best{0.278} & \second{0.242} & \best{0.278} & \best{0.240} & \second{0.281} & 0.253 & 0.291 & 0.287 & 0.342 & 0.370 & 0.408 & 0.263 & \second{0.281} & 0.260 & 0.299 & 0.261 & 0.299 & 0.258 & 0.300 \\
    \multirow{-4}{*}{\rotatebox{90}{\textbf{Weather}}} 
    & 720 & \second{0.311} & \best{0.327} & \best{0.309} & 0.332 & 0.312 & 0.334 & 0.317 & 0.337 & 0.362 & 0.393 & 0.420 & 0.421 & 0.314 & \second{0.331} & 0.325 & 0.345 & 0.325 & 0.346 & 0.329 & 0.348 \\
    \midrule

    & 96  & \second{0.131} & \second{0.224} & \best{0.129} & \best{0.221} & 0.141 & 0.240 & 0.135 & 0.233 & 0.140 & 0.237 & 0.226 & 0.341 & 0.139 & 0.231 & 0.139 & 0.239 & 0.147 & 0.253 & 0.142 & 0.247 \\
    & 192 & \best{0.148} & \best{0.237} & \best{0.148} & \second{0.238} & 0.156 & 0.256 & 0.155 & 0.253 & 0.165 & 0.259 & 0.220 & 0.336 & \second{0.153} & 0.245 & 0.155 & 0.250 & 0.159 & 0.256 & 0.164 & 0.273 \\
    & 336 & \best{0.164} & \best{0.256} & \second{0.165} & \second{0.257} & 0.172 & 0.267 & 0.169 & 0.267 & 0.190 & 0.286 & 0.224 & 0.337 & 0.169 & 0.262 & 0.171 & 0.265 & 0.169 & 0.270 & 0.171 & 0.260 \\
    \multirow{-4}{*}{\rotatebox{90}{\textbf{Electricity}}} 
    & 720 & \best{0.201} & \second{0.291} & \best{0.201} & \best{0.285} & 0.208 & 0.299 & \second{0.204} & 0.301 & 0.227 & 0.312 & 0.271 & 0.378 & 0.207 & 0.294 & 0.208 & 0.300 & 0.214 & 0.302 & 0.209 & 0.313 \\
    \bottomrule
  \end{tabular}
  \end{small}
  \end{threeparttable}
  }
\end{table*}

\end{document}